\documentclass[10pt]{article}
\usepackage[a4paper,margin=1.8cm,headsep=14pt]{geometry}
\usepackage[authoryear,round]{natbib}
\usepackage{lmodern}            
\usepackage{rhineai}            
\usepackage[utf8]{inputenc}
\usepackage{amsmath,amssymb}
\usepackage{booktabs}
\usepackage{array}
\usepackage{enumitem}
\usepackage{tikz}
\usetikzlibrary{arrows.meta,patterns,calc}
\usepackage[font=small,labelfont=bf,labelsep=period]{caption}
\usepackage{fancyhdr}
\usepackage{placeins}  
\usepackage{xurl}
\usepackage[colorlinks=true, linkcolor=rhineaiAccent, citecolor=rhineaiAccent, urlcolor=rhineaiAccent]{hyperref}
\usepackage[capitalise,noabbrev,nameinlink]{cleveref}

\newif\ifdraftnotes \draftnotestrue

\definecolor{paper}{HTML}{F7F5F1}
\definecolor{paper2}{HTML}{EFE3DB}
\definecolor{beige}{HTML}{E3CDC0}
\definecolor{ink}{HTML}{141414}
\definecolor{ink2}{HTML}{4A4440}
\definecolor{ink3}{HTML}{6B635B}
\definecolor{rulec}{HTML}{E3CDC0}
\definecolor{accent}{HTML}{960018}
\definecolor{posc}{HTML}{1F4E5F}
\definecolor{cellmid}{HTML}{ECE5DE}
\definecolor{mterra}{HTML}{2a78d6}
\definecolor{msol}{HTML}{1d4f9e}
\definecolor{mopus}{HTML}{b8461a}
\definecolor{mopus5}{HTML}{8a5a3c}
\definecolor{msonnet}{HTML}{eb6834}
\definecolor{mhaiku}{HTML}{f28c4e}
\definecolor{mfable}{HTML}{7c2d12}
\definecolor{mqwen}{HTML}{2f8f5b}
\definecolor{mluna}{HTML}{5c9ef0}
\definecolor{mglm}{HTML}{0a6136}
\definecolor{mhydrogen}{HTML}{8a1f6b}
\definecolor{mneon}{HTML}{c2358f}
\definecolor{mcarbon}{HTML}{e07ab8}
\definecolor{mastra62}{HTML}{00728f}
\definecolor{msol62}{HTML}{2fa3c2}
\definecolor{mdeepseek}{HTML}{5fae8a}
\definecolor{mflash}{HTML}{9c6b8c}
\definecolor{mastra6}{HTML}{5b4fa0}
\definecolor{mmistral}{HTML}{7a9a2e}

\newcommand{\uisf}{\fontfamily{phv}\selectfont}
\newcommand{\btw}{Bench the Weights}
\usepackage{fancyvrb}
\DefineVerbatimEnvironment{promptcode}{Verbatim}{fontsize=\footnotesize, xleftmargin=1em, formatcom=\color{ink2}}
\newcommand{\ph}[1]{\textcolor{accent}{\texttt{\{\{#1\}\}}}}
\newcommand{\prompthead}[2]{\par\ifnum#1<3\medskip\else\smallskip\fi\noindent{\uisf\bfseries #2}\par\smallskip}
\newcommand{\promptrule}{\par\smallskip\noindent{\color{rulec}\rule{\linewidth}{0.6pt}}\par\smallskip}
\newtcolorbox{promptbox}[1]{breakable, enhanced, colback=paper, colframe=rulec, boxrule=0.6pt, arc=2pt,
  left=8pt, right=8pt, top=6pt, bottom=6pt, title={#1}, fonttitle=\uisf\small\bfseries, coltitle=ink,
  colbacktitle=paper2, fontupper=\small, before upper={\raggedright\setlength{\parskip}{4pt}}}


\fancypagestyle{plain}{\fancyhf{}\fancyfoot[C]{\small\thepage}}

\newcommand{\swatch}[1]{\tikz[baseline=-0.6ex]\fill[#1] (0,-0.45ex) rectangle (1.1ex,0.65ex);}
\newcommand{\sw}[1]{\tikz[baseline=-0.25ex]\fill[#1] (0,-0.55ex) rectangle (1.5ex,0.95ex);}
\newcommand{\swdrop}{\tikz[baseline=-0.25ex]{\fill[accent!10!paper] (0,-0.55ex) rectangle (1.5ex,0.95ex);
  \fill[pattern=north east lines, pattern color=accent] (0,-0.55ex) rectangle (1.5ex,0.95ex);}}

\newcommand{\ecbody}[4]{\begin{tabular}{@{}c@{}}\textbf{#1}\\[-1pt]{\scriptsize\color{ink2}median #2 · #3}\\[-2pt]{\scriptsize\color{ink2}#4 passes}\end{tabular}}
\newcommand{\ec}[4]{{\setlength{\fboxsep}{2pt}\setlength{\fboxrule}{0.8pt}\fcolorbox{white}{white}{\ecbody{#1}{#2}{#3}{#4}}}}
\newcommand{\ecbest}[4]{{\setlength{\fboxsep}{2pt}\setlength{\fboxrule}{0.8pt}\fcolorbox{paper2}{paper2}{\ecbody{#1}{#2}{#3}{#4}}}}
\newcommand{\ecmain}[4]{{\setlength{\fboxsep}{2pt}\setlength{\fboxrule}{0.8pt}\fcolorbox{ink}{white}{\ecbody{#1}{#2}{#3}{#4}}}}
\newcommand{\ecbm}[4]{{\setlength{\fboxsep}{2pt}\setlength{\fboxrule}{0.8pt}\fcolorbox{ink}{paper2}{\ecbody{#1}{#2}{#3}{#4}}}}
\newcommand{\ecna}[1]{{\small\color{ink3}#1}}
\newcommand{\geqsev}{$\geq$75\%}
\newcommand{\hd}[2]{\begin{tabular}[b]{@{}c@{}}#1\\#2\end{tabular}}

\newcommand{\tval}[1]{{\setlength{\fboxsep}{1.2pt}\colorbox{paper2}{\strut\bfseries #1}}}
\newcommand{\tvalh}[1]{{\setlength{\fboxsep}{1.2pt}\colorbox{accent}{\strut\color{white}\bfseries #1}}}
\newcommand{\tst}[3]{{\color{ink2}\mbox{#1}} & \tval{#2}\ #3\\[3pt]}
\newcommand{\thot}[3]{{\color{accent}\bfseries\mbox{#1}} & \tvalh{#2}\ #3\\[3pt]}
\newcommand{\tnt}[1]{ & {\color{ink3}#1}\\[3pt]}
\newenvironment{trace}[1]{\begin{tabular}{@{}>{\raggedleft\arraybackslash}p{3.7em}@{\hspace{5pt}}!{\color{rulec}\vrule width 1.2pt}@{\hspace{6pt}}>{\raggedright\arraybackslash}p{#1}@{}}}{\end{tabular}}
\newcommand{\trhead}[3]{\noindent\swatch{#1}\,\textbf{#2}\hfill #3\par\vspace{3pt}{\color{rulec}\hrule height 0.6pt}\vspace{5pt}}

\setlist{itemsep=2pt, topsep=4pt}
\title{\btw: Can AI Agents Debug Neural Networks?}
\author{Robin Faro\rhineaiaff{1}\rhineainote{*}, Antonio Lopardo\rhineaiaff{1}\rhineainote{*}, Andrea Miele\rhineaiaff{1}\rhineainote{*}, Aurelien Lucchi\rhineaiaff{1}, Ilija Bogunovic\rhineaiaff{1}}
\rhineaiaffiliations{\rhineaiaff{1}University of Basel\quad\rhineainote{*}Equal contribution}

\rhineailabel{{\rhineaisf\color{rhineaiSoft}}}
\rhineaisite{}
\rhineaiprojectpage{\url{https://www.benchtheweights.com}}
\rhineaicode{\url{https://github.com/Rhine-AI-Lab/BenchTheWeights}}
\rhineailogo{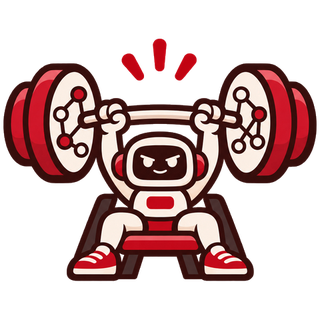}
\rhineailogowidth{2.3cm}

\begin{document}
\thispagestyle{plain}

\rhineaifrontmatter{Weight-space programming, the ability to read a network's weights and edit them directly
without gradients, is an emerging capability of coding agents. We introduce \btw{}, a benchmark that
tests whether AI agents can debug neural networks in weight space. Each of its 42 problems is a public
pretrained network, from seven domains and with up to 230 million parameters, in which a bug-injection
agent has placed two or three targeted bugs. The bugs touch at most 3\% of the weights, are hard to spot,
resist simple workarounds and can be reversed exactly. A solver agent receives the modified network with
25 minutes and 50 forward passes on a CPU, no gradients and no indication that anything has been changed, and
we score the share of the lost performance it recovers on a hidden test split. Across 19 frontier models
running in their providers' harnesses, the best agent, GPT-6.2 Astra in Codex, recovers 80\% on average
and solves 30 to 31 of the 42 problems, with at least 75\% recovery, while most models recover between 10\%
and 38\%. Trajectory
analysis shows that agents usually locate the bugs but often undo correct edits that do not yet improve
the score. Performance increases with more time, from 25\% to 83\% for Gemini 4 Neon at two hours, and has improved quickly across model releases.}

\begin{figure}[!h]
  \centering
  \includegraphics[width=0.8\linewidth]{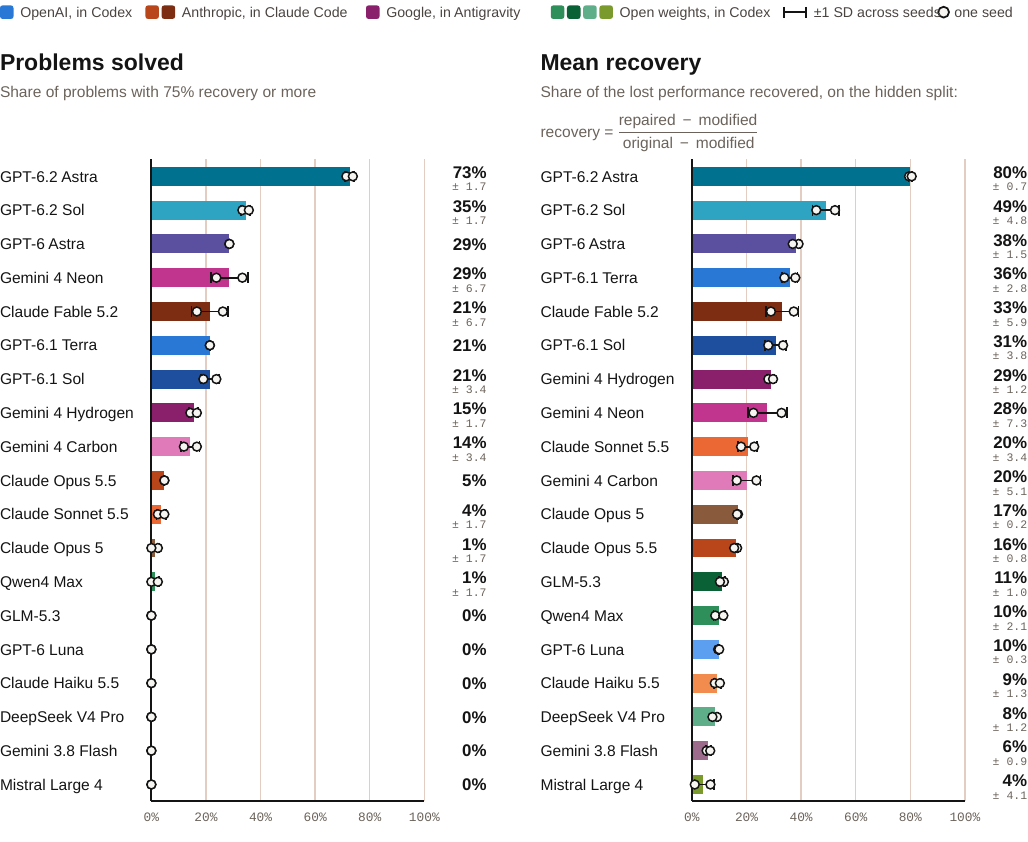}
  \caption{\textbf{Share of problems solved and mean recovery.} A seed is one trial on every problem,
  and only complete seeds count. Each bar is the mean over a model's seeds ($\pm$1 SD). Each chart ranks the models from best to worst.
  All models run at their highest reasoning effort.}
  \label{fig:solved}
\end{figure}

\section{Introduction}
\label{sec:intro}

Coding agents now handle a large part of everyday software work as well as a  growing part of machine-learning research engineering. A family of benchmarks tracks this progress on real GitHub issues \citep{jimenez2024swebench}, on ML experimentation and engineering tasks \citep{huang2024mlagentbench,chan2025mlebench,nathani2025mlgym}, on open-ended research engineering compared with human experts \citep{wijk2025rebench}, on the replication of published papers \citep{starace2025paperbench} and on the post-training of language models \citep{rank2026posttrainbench}, and the length of the tasks agents can complete has been growing steadily \citep{kwa2025horizon}. In all of these settings the agent works through code: it edits a script, runs it
and reads the output, while the network it trains is changed only by gradient descent techniques.

\paragraph{Weight-space programming.} A neural network is itself a program, written in weights. Programs can
be compiled into transformer weights by hand \citep{weiss2021rasp,lindner2023tracr}, interpretability
methods read algorithms back out of trained weights \citep{olah2020zoom,gupta2024interpbench}, and model
editing changes what a network knows through small, targeted weight updates
\citep{meng2022rome,meng2023memit,ilharco2023task}. Reading a network's parameters and editing them
directly, without a learning algorithm or large amounts of data, has recently been dubbed weight-space
programming, and \citet{takehi2026sparks} show that frontier models working as coding agents can
already inspect and modify network weights in this way. Concurrent work has started to evaluate this
ability more broadly, on reading, fixing and writing tasks \citep{marasini2026weightbench} and on the
repair of vision-language models \citep{aluguvelly2026vlm}.

\paragraph{Debugging as a test case.} When a model breaks or behaves unexpectedly, we often have little
idea what has gone wrong without retraining it or testing it extensively. An agent that can inspect, diagnose
and repair a network directly could help interpret the role of individual layers and weights, explain
failed training runs and fine-tunes, and make networks easier to maintain. Debugging also has a long
tradition of evaluation on known bugs in software engineering, where faults are either injected on
purpose \citep{jia2011mutation} or collected from real projects
\citep{just2014defects4j,jimenez2024swebench}. A known bug provides a ground truth, so one can check
whether a tool finds and fixes the right defect, and not only whether a test starts passing. Fault injection
has long been used to evaluate testing and repair tools for neural networks
\citep{ma2018deepmutation,humbatova2021deepcrime}, and benchmarks with a planted answer ask agents to
describe a model or uncover a hidden behaviour \citep{schwettmann2023find,marks2025auditing}. We use this
protocol to study whether general-purpose agents can locate bugs and reverse them directly in the weights,
without a predefined repair algorithm.

\paragraph{Our benchmark.} We introduce \textbf{\btw{}}, a benchmark of 42 debugging problems in weight space. Each
problem is a public pretrained network, from seven domains and with 1 to 230 million parameters, in
which a bug-injection agent has placed two or three targeted bugs touching at most 3\% of the weights
(\cref{sec:benchmark}). Every problem passes three tests: the bugs are hard to spot from weight
statistics, they resist simple workarounds such as refitting a layer, and they can be reversed exactly,
as shown by a reference repair. A solver agent receives the modified network with 25 minutes and 50
forward passes, and a separate verifier scores its signed edits on a hidden test split. Because we know
which weights are modified, we can check whether an agent reverses the bugs and analyse each step of its
trajectory against the answer key.

\paragraph{Our findings.} We evaluate 19 frontier models from seven labs, each running as a coding agent in
its provider's harness, or Codex for open-weight models (\cref{sec:results}). GPT-6.2 Astra recovers 80\% of the lost performance on
average and solves 30 to 31 of the 42 problems, about twice as many as the second-best model,
while six models follow between 28\% and 38\% (\cref{fig:solved}). Repairs are mostly all or nothing, and
the leading model stands out by solving more problems. Capability on the task has risen
quickly: no model released before September 2026 solves more than one problem, and most of the
progress comes from models released in November. The trajectories show that most agents locate the bugs, and that most failures happen
afterwards, when an agent undoes a correct edit because the score has not moved yet
(\cref{sec:behavior}). The strongest runs keep such edits, compare the network before and after their edits on
the same inputs, and use the whole time budget. Time is the most important setting: with two hours instead
of 25 minutes, Gemini 4 Neon goes from 25\% to 83\% recovery, whereas reasoning effort and the harness
have no consistent effect (\cref{sec:ablations}).

\section{\btw}
\label{sec:benchmark}

\textbf{Task definition.} Given a pretrained network whose weights have been modified, the agent must
restore the network's performance on its original task by editing its weights. The agent may inspect
any weight, run forward passes on development examples or on inputs of its own, write and run arbitrary
Python code in its sandbox, and edit up to 3\% of the parameters. The agent may not compute gradients,
train on development data, access the internet beyond its own model API, or read the original weights.
It is not told that the network has been modified.

Each problem goes through the same pipeline (\cref{fig:pipeline}). We collect a pretrained network, a
bug-injection agent modifies its weights, the modified network is handed to a solver agent, and a
verifier scores the solver's repair. The rest of this section describes the networks, the bugs and the
solver's environment.

\subsection{Networks}
\label{sec:networks}

\begin{figure}[htbp]
  \centering
  \includegraphics[width=0.92\linewidth]{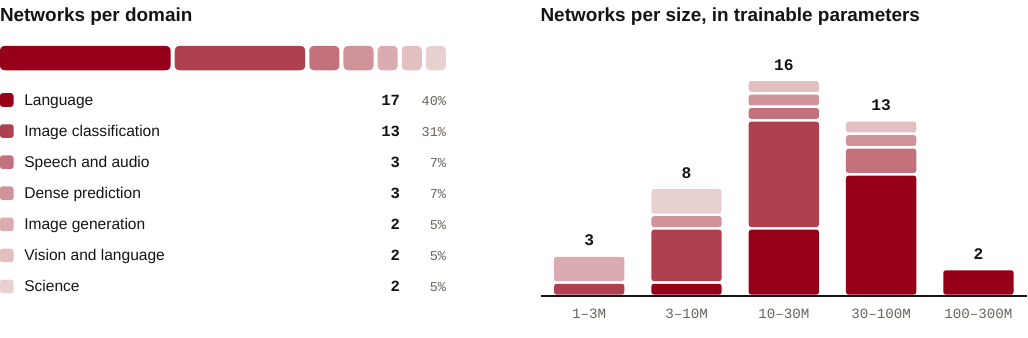}
  \caption{\textbf{Domains and sizes of the 42 networks.} Most have between 10 and 100 million
  parameters. All of the image classifiers have fewer than 30 million, while the language models reach
  230 million. Colours mark the domain, as on the left.}
  \label{fig:domains}
\end{figure}

We start from public pretrained networks that come with an evaluation setup and a task metric. Language
and image classification make up 30 of the 42. The rest cover speech and audio, dense prediction such as
segmentation and depth estimation, image generation, vision and language, and science. Sizes run from
about 1 million to 230 million trainable parameters, with a median of 23 million (\cref{fig:domains}).

\subsection{Injecting Bugs}
\label{sec:bugs}

\begin{figure}[htbp]
  \centering
  \resizebox{\linewidth}{!}{
\begin{tikzpicture}[
    font=\uisf\scriptsize, x=1cm, y=1cm,
    box/.style={draw=ink3, fill=paper, rounded corners=2.5pt, line width=0.5pt},
    key/.style={line width=0.9pt},
    arr/.style={-{Latex[length=4.5pt,width=3.5pt]}, line width=0.6pt, color=ink2},
    redarr/.style={-{Latex[length=4.5pt,width=3.5pt]}, line width=0.6pt, color=accent},
    t1/.style={anchor=north west, inner sep=0pt, font=\uisf\footnotesize\bfseries, color=ink},
    t2/.style={anchor=north west, inner sep=0pt, color=ink2, align=left},
    t3/.style={anchor=north west, inner sep=0pt, font=\ttfamily\scriptsize, color=ink2, align=left},
    lb/.style={inner sep=1.5pt, color=ink3},
    stp/.style={anchor=west, inner sep=0pt, font=\uisf\scriptsize\bfseries},
  ]
  \node[stp, color=ink3]   at (0,0)    {1 · Collect};
  \node[stp, color=accent] at (3.2,0) {2 · Inject bugs};
  \node[stp, color=accent] at (8.95,0) {3 · Filter};
  \node[stp, color=posc]   at (11.85,0){4 · Hand over};
  \node[stp, color=posc]   at (15.35,0){5 · Repair and score};

  \draw[box, fill=beige!45!paper] (0,-0.55) rectangle (2.7,-2.45);
  \node[t1, font=\uisf\scriptsize\bfseries] at (0.2,-0.75) {Pretrained model};
  \node[t2] at (0.2,-1.2) {public, works well};
  \node[t3] at (0.2,-1.65) {clean score 71\%};
  \draw[arr] (2.7,-1.5) -- (3.2,-1.5);

  \draw[box, key, draw=accent, fill=accent!5!paper] (3.2,-0.35) rectangle (8.35,-4.0);
  \node[t1, color=accent] at (3.4,-0.55) {Bug-injection agent};
  \draw[box, draw=accent!45!paper] (3.4,-1.0) rectangle (5.7,-1.95);
  \node[t1, font=\uisf\scriptsize\bfseries] at (3.55,-1.15) {Design bugs};
  \node[t2] at (3.55,-1.55) {2–3 bugs, $\leq$\,3\%};
  \draw[box, draw=accent!45!paper] (5.95,-1.0) rectangle (8.15,-1.95);
  \node[t1, font=\uisf\scriptsize\bfseries] at (6.1,-1.15) {Fix them};
  \node[t2] at (6.1,-1.55) {as a solver would};
  \draw[arr] (5.7,-1.45) -- (5.95,-1.45);
  \draw[arr] (7.05,-1.95) -- (7.05,-2.6);
  \draw[box, draw=accent!45!paper] (3.4,-2.6) rectangle (8.15,-3.8);
  \node[t1, font=\uisf\scriptsize\bfseries] at (3.55,-2.75) {Three tests};
  \node[t2] at (3.55,-3.15) {hard to spot, no shortcuts,\\exactly repairable};
  \draw[redarr] (4.35,-2.6) -- (4.35,-1.95);
  \node[anchor=west, inner sep=0pt, color=accent] at (4.47,-2.28) {any test fails: rework};
  \draw[arr] (8.35,-1.5) -- (8.95,-1.5);
  \node[lb, above, inner sep=1pt, font=\uisf\tiny] at (8.65,-1.5) {pass};

  \draw[box, draw=accent!55!paper, fill=accent!4!paper] (8.95,-0.55) rectangle (11.25,-3.2);
  \node[t1] at (9.15,-0.75) {Filter};
  \node[t2] at (9.15,-1.2) {drop it if no\\design passes,\\or if simple\\tricks solve it};
  \node[t3] at (9.15,-2.45) {71 networks\\$\to$ 50 $\to$ 42};
  \draw[arr] (11.25,-1.5) -- (11.85,-1.5);
  \draw[redarr] (10.1,-3.2) -- (10.1,-3.75);
  \node[anchor=north, inner sep=1pt, color=accent] at (10.1,-3.8) {dropped};

  \draw[box, draw=posc!55!paper, fill=posc!4!paper] (11.85,-0.55) rectangle (14.85,-3.2);
  \node[t1] at (12.05,-0.75) {Sandbox};
  \node[t2] at (12.05,-1.2) {task prompt and tools\\modified weights\\examples with answers\\current, reference score};
  \node[t2, font=\uisf\scriptsize\itshape] at (12.05,-2.75) {not told what was done};
  \draw[arr] (14.85,-1.5) -- (15.35,-1.5);

  \draw[box, key, draw=posc, fill=posc!6!paper] (15.35,-0.35) rectangle (20.25,-3.85);
  \node[t1, color=posc] at (15.55,-0.55) {Solver: model + harness};
  \draw[box, draw=posc!45!paper] (15.55,-0.95) rectangle (20.05,-1.55);
  \node[t2, anchor=west] at (15.7,-1.25) {Python kernel in a container};
  \draw[arr] (17.2,-1.55) -- (17.2,-2.05);
  \node[lb, anchor=west] at (17.3,-1.8) {every read, run, edit};
  \draw[box, draw=posc!45!paper] (15.55,-2.05) rectangle (20.05,-3.65);
  \node[t1, font=\uisf\scriptsize\bfseries] at (15.7,-2.2) {Backend};
  \node[t3] at (15.7,-2.6) {25 min · 50 forward passes\\3\% of weights · no gradients};
  \node[t2] at (15.7,-3.25) {signs each edit it accepts};
  \draw[arr] (17.2,-3.85) -- (17.2,-4.45);
  \node[lb, anchor=west] at (17.3,-4.15) {signed edits};

  \draw[box, draw=posc!55!paper, fill=posc!4!paper] (15.35,-4.45) rectangle (20.25,-5.7);
  \node[t1] at (15.55,-4.65) {Verifier};
  \node[t2] at (15.55,-5.05) {applies the signed edits\\recovery on the hidden split};

  \node[anchor=north west, inner sep=0pt, color=accent, align=left] at (0,-4.75) {Never reaches\\the solver};
  \draw[arr] (5.775,-4.0) -- (5.775,-4.45);
  \draw[box, dashed, draw=accent!55!paper, fill=accent!4!paper2] (3.2,-4.45) rectangle (8.35,-5.7);
  \node[t1] at (3.4,-4.65) {Answer key and reference repair};
  \node[t2] at (3.4,-5.05) {the injection agent's own blind fix,\\which must reach about 100\%};
  \draw[box, dashed, draw=accent!55!paper, fill=accent!4!paper2] (11.85,-4.45) rectangle (14.85,-5.7);
  \node[t1] at (12.05,-4.65) {Hidden test split};
  \node[t2] at (12.05,-5.05) {inputs the solver\\never sees};
  \draw[arr] (14.85,-5.075) -- (15.35,-5.075);
\end{tikzpicture}}
  \caption{\textbf{From a pretrained model to a scored repair.} Problems that fail the filter are
  dropped. The harness is Codex, Claude Code or Antigravity, depending on the model. Dashed boxes stay
  with us: the solver never sees the answer key, the reference repair or the hidden test split.}
  \label{fig:pipeline}
\end{figure}
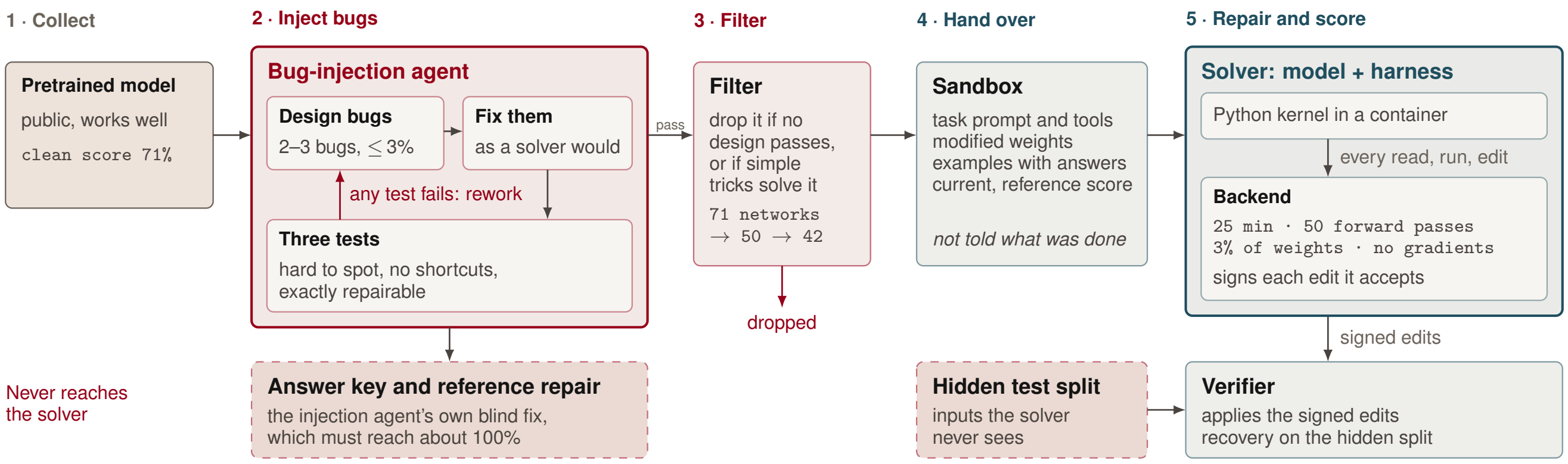

For each network, a bug-injection agent designs two or three small, targeted bugs that touch at most
3\% of the weights (\cref{fig:pipeline}). The bugs delete nothing: they move, reorder or re-sign what the
network has learned, so the original can always be restored. The agent tries each design the way a
solver would, with the solver's tools and budget, and reworks it until it passes three tests:
\begin{enumerate}[leftmargin=1.5em]
  \item \textbf{Hard to spot.} The modified weights look like normal trained weights. The bugs show
  only when the network runs.
  \item \textbf{No shortcuts.} Fixes that raise the score without explaining the bugs, such as deleting
  the modified part, averaging it with its neighbours or refitting a layer on the dev examples, recover
  less than 50\%. Undoing any one bug alone recovers less than 20\%.
  \item \textbf{Exactly repairable.} The agent's own repair, found with only the solver's tools and
  budget, reaches about 100\% of the original performance.
\end{enumerate}
The agent runs headless in Claude Code at its highest reasoning effort, mostly with Claude Opus 5.5,
Claude Fable 5.1 or GPT-6 Astra. It returns the modified weights, a repair script and a rationale. Its
prompt is in \cref{app:prompt-injection}.

\textbf{Filtering.} From 71 candidate networks, the agent built 50 problems that pass the three tests.
The other attempts found no design that met the requirements, were aborted, or moved on to another
network. We then dropped 8 of the 50 problems because simple tricks could solve them, which leaves the
42 in this version.

\begin{figure}[htbp]
  \centering
  \includegraphics[width=0.95\linewidth]{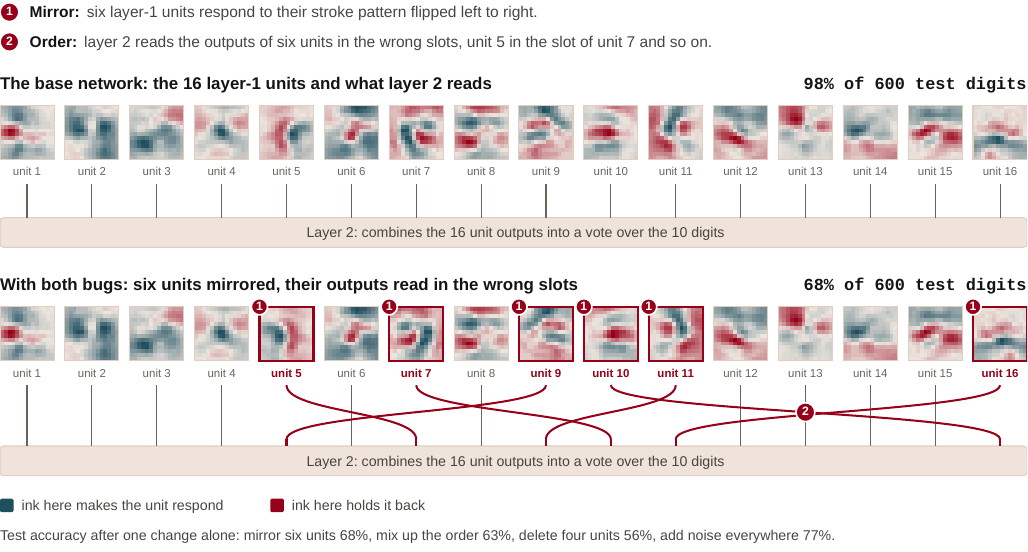}
  \caption{\textbf{Two bugs in a small digit classifier.} The 16 first-layer units of the network and the
  wiring from their outputs into the second layer, before and after a combination of bugs that passes all
  three tests. Bug~1 mirrors six units; bug~2 is independent of it and reroutes the outputs of those six
  units into the wrong input slots of the second layer.}
  \label{fig:toy}
\end{figure}

\Cref{fig:toy} shows an example of such bugs in a toy network, a two-layer classifier of
handwritten digits. Mirroring six first-layer units and mixing up the order in which the second layer
reads them gives a problem that passes all three tests. Either bug undone alone recovers almost nothing,
and refitting the second layer on 50 labelled examples recovers about a third of the loss.

\subsection{Solver Setup}
\label{sec:solver}

\textbf{Agents and harnesses.} We run 19 models as coding agents. Six
OpenAI models run in Codex, five Anthropic models in Claude Code, four Google models in Antigravity, and
four open-weight models, Qwen, GLM, DeepSeek and Mistral, in Codex. Every model runs each problem on two seeds,
at its highest reasoning effort (\cref{tab:models}).
 
\textbf{Inputs and budget.} The solver agent gets the modified network, its current score and the
original network's score, presented as the score of another network with the same architecture. It is
not told that anything has been changed. It works in a sandboxed Python environment, and every read, forward
pass and edit goes through a backend that tracks its budget and signs the edits it accepts. Every network
runs on a CPU with 4\,GB of memory. When time runs out, a separate verifier applies the signed edits and
scores the network on a hidden test split. We also ask each agent to start every code cell with a
one-line hypothesis, such as ``the outputs of the third block look off'', so we can see what it believed
at each step. Its full instruction is in \cref{app:prompt-solver}.

\begin{center}\small
\begin{tabular}{@{}l l l@{}}
  \toprule
  Working time   & \textbf{25 min}    & The clock pauses while a forward pass runs \\
  Forward passes & \textbf{50}        & Each on a batch of dev examples or inputs of the agent's own \\
  Edits          & \textbf{3\%}       & Of all parameters, counted over every edit in the trial \\
  Gradients      & \textbf{none}      & No optimiser and no training on dev data \\
  Network        & \textbf{sandboxed} & Only the agent's own model API is reachable \\
  \bottomrule
\end{tabular}
\end{center}

\textbf{Scoring.} Each problem keeps its own metric, such as accuracy, mIoU, word error rate or PSNR. To
compare problems, we report recovery, the share of the lost performance a repair wins back on the hidden
split. A solver that brings a modified classifier from 12\% back to 49\%, when the original scores 71\%,
has a recovery of (49~$-$~12)~/~(71~$-$~12), or 63\%:
\begin{equation}
  \text{recovery} = \frac{\text{repaired} - \text{modified}}{\text{original} - \text{modified}} .
  \label{eq:recovery}
\end{equation}
\textbf{What counts as a fix.} We score the function of the repaired network, not its weights. A repair
counts if the network scores as well as the original on the hidden split, whether or not every modified
weight returns to its original value. This matters because many of our bugs reorder units, and a
network computes the same function with its units in a different order. The two or three bugs of a
problem are scored together, through the recovery of the whole network, and not one at a time. We call
a problem \emph{solved} when a trial reaches at least 75\% recovery, and \emph{untouched} when it
recovers less than 5\%; we use ``solved'' and ``repaired'' in this sense throughout. Recovery can exceed
100\% when the repaired network scores above the original, and is negative when a repair leaves the
network worse than the modified one. \Cref{sec:behavior} uses the answer key separately to tell whether
a trial found a bug, that is, edited one of the modified weight tensors or named it in a hypothesis.

\textbf{Evaluation protocol.} Every model runs each problem twice, at its highest reasoning effort, with
the budget above. A seed is one trial on every problem, and only complete seeds count. We report two
metrics per model, mean recovery over the 42 problems and the share of problems solved, each as mean
$\pm$ standard deviation over seeds. The evaluated models, their providers, harnesses and release dates are
listed in \cref{tab:models}.

\textbf{Compute and cost.} The main runs comprise about 1,600 trials, 84 per model, and about 600 hours
of agent working time. The networks run on CPUs only, so evaluating a model on \btw{} needs no GPU. At
API list prices, the main runs cost about \$4,000, estimated from each model's median cost per trial
(\cref{fig:cost}). Building the 42 problems took about 150 hours of
agent time and about \$700 at the same prices, a median of 2.6 hours and \$13 per problem (from \$4 to
\$59). With the discarded attempts, this rises to about 250 hours and \$1,000 to \$1,200, or about \$25
per problem kept. We did not record GPU hours for the bug-injection runs.

\section{Experimental Results}
\label{sec:results}

We organise the results around five questions: which models repair the most (\textbf{Q1},
\cref{sec:main}), how quickly this ability is improving (\textbf{Q2}, \cref{sec:evolved}), whether the
agents find the bugs at all (\textbf{Q3}, \cref{sec:finding}), what separates the best runs from the
rest (\textbf{Q4}, \cref{sec:hypotheses,sec:persistence}), and how much the results depend on time,
reasoning effort and harness (\textbf{Q5}, \cref{sec:ablations}). This section answers Q1 and Q2,
\cref{sec:ablations} answers Q5, and \cref{sec:behavior} looks at what the agents do to answer Q3 and
Q4.

\begin{table}[htbp]
  \caption{\textbf{Leaderboard.} Mean recovery and share of problems solved (75\% recovery or more), in
  percent, as mean $\pm$ SD over each model's two complete seeds, as in \cref{fig:solved}. The last three
  columns give the mean recovery per domain: language (17 networks), image classification (13) and the
  other five domains (12). The best value in each column is in bold. Per-domain values are averaged over
  problems without clipping, so a repair that leaves a network worse than its modified version counts as
  negative. All models run at their highest reasoning effort, with 25 minutes and 50 forward passes per
  trial.}
  \label{tab:leaderboard}
  \centering\small
  \setlength{\tabcolsep}{5pt}
  \begin{tabular}{@{}r l l r r r r r@{}}
    \toprule
    & & & & & \multicolumn{3}{c}{Recovery by domain} \\
    \cmidrule(l){6-8}
    Rank & Model & Harness & Recovery & Solved & Language & Image cls. & Other \\
    \midrule
1 & \swatch{mastra62}\,GPT-6.2 Astra & {\footnotesize\color{ink3}Codex} & \textbf{80.0}\,{\scriptsize$\pm$\,0.7} & \textbf{72.6}\,{\scriptsize$\pm$\,1.7} & \textbf{80.7} & \textbf{96.5} & \textbf{61.1} \\
2 & \swatch{msol62}\,GPT-6.2 Sol & {\footnotesize\color{ink3}Codex} & 49.0\,{\scriptsize$\pm$\,4.8} & 34.5\,{\scriptsize$\pm$\,1.7} & 46.5 & 59.8 & 40.7 \\
3 & \swatch{mastra6}\,GPT-6 Astra & {\footnotesize\color{ink3}Codex} & 38.0\,{\scriptsize$\pm$\,1.5} & 28.6\,{\scriptsize$\pm$\,0.0} & 43.0 & 37.6 & 31.3 \\
4 & \swatch{mterra}\,GPT-6.1 Terra & {\footnotesize\color{ink3}Codex} & 35.9\,{\scriptsize$\pm$\,2.8} & 21.4\,{\scriptsize$\pm$\,0.0} & 35.6 & 44.1 & 27.5 \\
5 & \swatch{mfable}\,Claude Fable 5.2 & {\footnotesize\color{ink3}Claude Code} & 33.1\,{\scriptsize$\pm$\,5.9} & 21.4\,{\scriptsize$\pm$\,6.7} & 31.1 & 43.4 & 24.8 \\
6 & \swatch{msol}\,GPT-6.1 Sol & {\footnotesize\color{ink3}Codex} & 30.7\,{\scriptsize$\pm$\,3.8} & 21.4\,{\scriptsize$\pm$\,3.4} & 29.4 & 39.9 & 22.4 \\
7 & \swatch{mhydrogen}\,Gemini 4 Hydrogen & {\footnotesize\color{ink3}Antigravity} & 28.8\,{\scriptsize$\pm$\,1.2} & 15.5\,{\scriptsize$\pm$\,1.7} & 28.0 & 35.2 & 23.1 \\
8 & \swatch{mneon}\,Gemini 4 Neon & {\footnotesize\color{ink3}Antigravity} & 27.7\,{\scriptsize$\pm$\,7.3} & 28.6\,{\scriptsize$\pm$\,6.7} & 38.7 & 49.2 & -11.3 \\
9 & \swatch{msonnet}\,Claude Sonnet 5.5 & {\footnotesize\color{ink3}Claude Code} & 20.4\,{\scriptsize$\pm$\,3.4} & 3.6\,{\scriptsize$\pm$\,1.7} & 20.6 & 19.6 & 21.0 \\
10 & \swatch{mcarbon}\,Gemini 4 Carbon & {\footnotesize\color{ink3}Antigravity} & 20.0\,{\scriptsize$\pm$\,5.1} & 14.3\,{\scriptsize$\pm$\,3.4} & 21.6 & 19.8 & 18.0 \\
11 & \swatch{mopus5}\,Claude Opus 5 & {\footnotesize\color{ink3}Claude Code} & 16.7\,{\scriptsize$\pm$\,0.2} & 1.2\,{\scriptsize$\pm$\,1.7} & 20.5 & 9.4 & 19.2 \\
12 & \swatch{mopus}\,Claude Opus 5.5 & {\footnotesize\color{ink3}Claude Code} & 16.0\,{\scriptsize$\pm$\,0.8} & 4.8\,{\scriptsize$\pm$\,0.0} & 22.7 & 6.5 & 16.9 \\
13 & \swatch{mglm}\,GLM-5.3 & {\footnotesize\color{ink3}Codex} & 11.0\,{\scriptsize$\pm$\,1.0} & 0.0\,{\scriptsize$\pm$\,0.0} & 11.4 & 8.0 & 13.7 \\
14 & \swatch{mqwen}\,Qwen4 Max & {\footnotesize\color{ink3}Codex} & 10.0\,{\scriptsize$\pm$\,2.1} & 1.2\,{\scriptsize$\pm$\,1.7} & 14.7 & 8.6 & 4.9 \\
15 & \swatch{mluna}\,GPT-6 Luna & {\footnotesize\color{ink3}Codex} & 9.8\,{\scriptsize$\pm$\,0.3} & 0.0\,{\scriptsize$\pm$\,0.0} & 11.8 & 8.4 & 8.4 \\
16 & \swatch{mhaiku}\,Claude Haiku 5.5 & {\footnotesize\color{ink3}Claude Code} & 9.3\,{\scriptsize$\pm$\,1.3} & 0.0\,{\scriptsize$\pm$\,0.0} & 14.0 & 6.7 & 5.6 \\
17 & \swatch{mdeepseek}\,DeepSeek V4 Pro & {\footnotesize\color{ink3}Codex} & 8.3\,{\scriptsize$\pm$\,1.2} & 0.0\,{\scriptsize$\pm$\,0.0} & 8.8 & 7.6 & 8.4 \\
18 & \swatch{mflash}\,Gemini 3.8 Flash & {\footnotesize\color{ink3}Antigravity} & 6.0\,{\scriptsize$\pm$\,0.9} & 0.0\,{\scriptsize$\pm$\,0.0} & 7.3 & 4.2 & 6.3 \\
19 & \swatch{mmistral}\,Mistral Large 4 & {\footnotesize\color{ink3}Codex} & 3.9\,{\scriptsize$\pm$\,4.1} & 0.0\,{\scriptsize$\pm$\,0.0} & 6.4 & 3.5 & 0.8 \\
    \bottomrule
  \end{tabular}
\end{table}

\subsection{Who Repairs the Most?}
\label{sec:main}

\paragraph{Overall performance.} GPT-6.2 Astra leads by a wide margin (\cref{fig:solved},
\cref{tab:leaderboard}). It recovers 80\% of the lost performance on average and solves 30 to 31 of the 42
problems, depending on the seed. GPT-6.2 Sol is second at 49\%. Behind them we observe a cluster of six
models between 28\% and 38\%: GPT-6 Astra, GPT-6.1 Terra and Sol, Claude Fable 5.2, and Gemini 4 Hydrogen
and Neon. Sonnet, Carbon and both Opus models recover 16 to 20\%, and the open-weight models, GPT-6 Luna,
Haiku and Gemini 3.8 Flash recover about a tenth or less.

\paragraph{Solved problems and mean recovery.} The two metrics mostly agree, with two exceptions. Neon solves
as many problems as GPT-6 Astra, the most in its group, while recovering the least on average, and Sonnet
recovers as much as Carbon while solving almost nothing.

\begin{figure}[htbp]
  \centering
  \includegraphics[width=\linewidth]{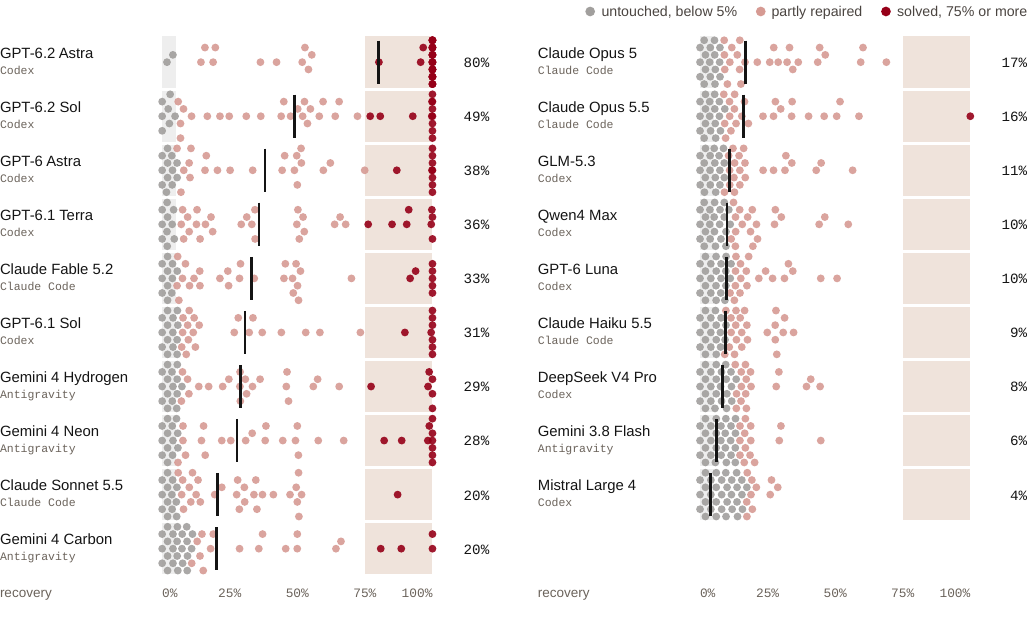}
  \caption{\textbf{Recovery on each problem.} One dot per problem, averaged over a model's complete
  seeds; the vertical line is the mean and the shaded band is the solved range. The dots are coloured
  with the scale above: untouched below 5\%, partly repaired, and solved at 75\% or more.}
  \label{fig:strip}
\end{figure}

\paragraph{Per-problem variation.} The mean hides how uneven the results are (\cref{fig:strip}). Every model
except GPT-6.2 Astra leaves most problems partly repaired or untouched, and even Astra leaves 1 to 3
untouched per seed. Of the 798 results in \cref{fig:strip}, 78 recover 95\% or more, only 34 land between
60\% and 95\%, and the partial repairs have a median of 21\%. This distribution is expected from the design
of the problems: the bugs cover for each other, so a workaround or a fix for a single bug recovers little,
while an agent that identifies all of them can reverse them exactly.

\paragraph{Problems solved by a single model.} GPT-6.2 Astra's lead comes from solving more problems.
Averaged over its seeds, it solves 13 problems that no other model solves, and misses only
one of the 18 that some other model solves. Eleven problems are solved by no model, which leaves room
for future models on the current version of the benchmark.

\subsection{How Capability Has Evolved}
\label{sec:evolved}

\begin{figure}[htbp]
  \centering
  \includegraphics[width=\linewidth]{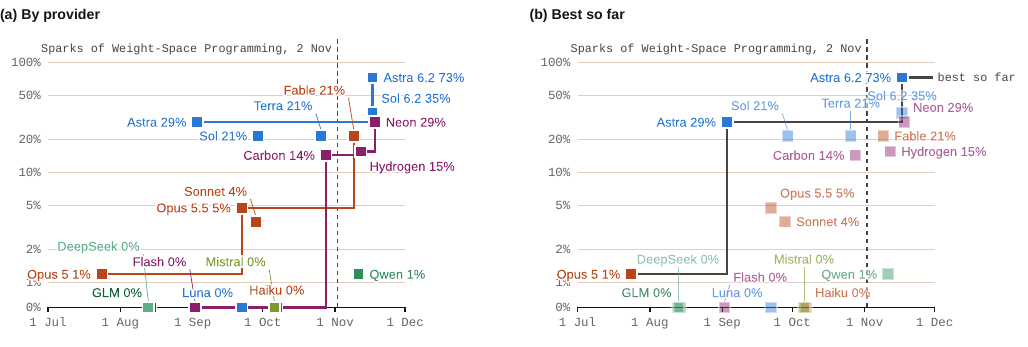}
  \caption{\textbf{Problems solved against release date.} The share of problems with 75\% recovery or
  more, as in \cref{fig:solved}, on a logarithmic axis from 1\% to 100\%; models that solve no problem
  sit on the separate 0 row at the bottom. \emph{By provider}: each provider's line joins only the models
  that beat that provider's best so far, staying flat between them and then stepping up; models that do
  not beat it are points off the line. \emph{Best so far}: the line steps through the models that are
  the best of any provider at release; the others are faded. The dashed line marks the publication of
  Sparks of Weight-Space Programming (2 November 2026).}
  \label{fig:timeline}
\end{figure}

\paragraph{Progress across releases.} The 19 models span four months of releases, so the same results
also show how fast weight-space programming has improved (\cref{fig:timeline}). The progress is not smooth. No model released before September
solves more than one of the 42 problems. GPT-6 Astra, released on 3 September, solves 29\%, and
no model released in the following two and a half months beats it. GPT-6.2 Astra, released on 17 November,
then reaches 73\%, two and a half times the previous best.

\paragraph{Timing.} Most of this progress comes from models released in November. Sparks of Weight-Space Programming
\citep{takehi2026sparks} dates from 2 November, and every provider that has released a model since then
beats its earlier best. OpenAI goes from 29\% to 73\% with GPT-6.2 Astra, Google from 14\% to 29\% with Gemini 4 Neon, and
Anthropic from 5\% to 21\% with Claude Fable 5.2. The timing does not establish cause and effect, since the
development of these models very likely predates the post. The results do suggest that models are
improving quickly at editing network weights directly, and the pace may indicate that frontier labs
already train on similar tasks. With 42 problems and two seeds per model, the steps between neighbouring models are
rough.

\section{Ablation Studies}
\label{sec:ablations}

So far every model runs with the same settings: 25 minutes, 50 forward passes and its highest reasoning
effort. In this section, we ablate how the choice of each setting changes the performance on the task. These runs being expensive, we use a subset of 9 problems out of the full set of 42 problems: three each where the mean recovery across models is low, medium and high, picking in each group the problems with the smallest spread in recovery. For each model, we run two trials per problem and setting.

\subsection{Effect of the Time Limit}
\label{sec:time}

\begin{figure}[htbp]
  \centering
  \includegraphics[width=\linewidth]{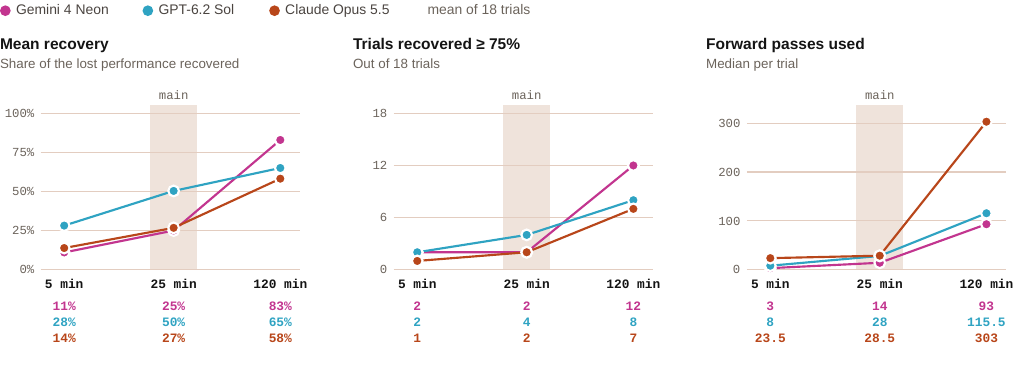}
  \caption{\textbf{Gemini 4 Neon, GPT-6.2 Sol and Claude Opus 5.5 with 5, 25 and 120 minutes.} Nine
  problems, two trials each, at xhigh effort. The shaded column is the setting of the main results. The
  25-minute runs allow 50 forward passes and the 120-minute runs 1000; Sol's and Opus's 5-minute runs
  also allow 1000, and Neon's have no practical limit. The time axis is logarithmic. The medians are in
  \cref{tab:timebudget}.}
  \label{fig:timebudget}
\end{figure}

The first setting that we inspect is the time limit, fixed at 25 minutes for the benchmark. We want to see if with more time, the agents are capable of recovering more performance. To investigate this, we give Gemini 4 Neon, GPT-6.2 Sol and Claude Opus 5.5 either 5, 25 or 120 minutes on the nine problems
(\cref{fig:timebudget}). Five minutes leaves little time beyond understanding the task, and Neon runs a
median of only 3 forward passes for 11\% mean recovery. At 120 minutes its mean recovery rises from 25\%
to 83\%, with 12 of 18 trials at 75\% or more, against 2 at 25 minutes. Forward passes are not the
bottleneck. Even with 1,000 available, the median trial uses 93.

Sol starts stronger but improves more slowly. Its mean recovery goes from 28\% at 5 minutes to 50\% at
25 and 65\% at 120, and its trials at 75\% or more from 2 to 4 to 8 of 18. Time is again the
constraint: at 120 minutes the median trial uses 116 forward passes, and none more than 365. With enough
time, Neon overtakes Sol, 83\% against 65\%.

Claude Opus 5.5 follows the same curve from a lower start. Its mean recovery goes from 14\% to 27\% to
58\%, and its trials at 75\% or more from 1 to 2 to 7 of 18. With two hours it uses far more forward
passes than the others, a median of 303, and still stops on its own, after a median of 78 minutes and
never later than 98.

With more time, agents often recover all of the lost performance. Neon goes from 15\% to 100\% on a
language task and from 0\% to 100\% on an image classifier. An earlier, smaller experiment with Claude
Opus, Sonnet and Haiku, at 120 minutes and 300 forward passes on five problems, shows the same pattern.
Opus goes from 40\% to 99\% on a language task and from 18\% to 100\% on an image classifier, mostly
after the first 30 minutes. The results on the full set therefore measure how fast an agent finds and
fixes the bugs, as well as whether it can.

\subsection{Reasoning Effort}
\label{sec:thinking}

In the full runs, we fix another parameter: the reasoning effort, which trades cost for quality. To see
whether it matters under a time limit, we run seven models at lower efforts on the nine problems, with
the standard 25 minutes and 50 forward passes (\cref{tab:effort}). The effect is not consistent across
models.

\begin{table}[htbp]
  \caption{\textbf{GPT-6.2 Sol, GPT-6.1 Terra, Claude Opus 5.5 and Sonnet 5.5, and Gemini 4 Hydrogen,
  Neon and Carbon by reasoning effort.} Each cell gives the mean recovery, its median, the trials that
  recover at least 75\% and the median forward passes. The shaded cell is each model's best effort; the
  outlined one is its setting in the main results. Nine problems, two trials each (four for Opus at
  medium), with 25 minutes and 50 forward passes. The bottom row checks Neon's best setting on all 42
  problems, two trials each: the xhigh cell is the main results, and the medium cell adds 66 new trials
  to the 18 above.}
  \label{tab:effort}
  \centering\small
  \setlength{\tabcolsep}{3pt}
  \begin{tabular}{@{}l c c c c@{}}
    \toprule
    & low & medium & high & xhigh \\
    \midrule
    Sol 6.2    & \ec{49\%}{40\%}{5/18 \geqsev}{33}   & \ec{35\%}{30\%}{3/18 \geqsev}{22}     & \ec{46\%}{39\%}{4/18 \geqsev}{31.5}     & \ecbm{50\%}{49\%}{4/18 \geqsev}{28} \\
    Terra 6.1  & \ec{26\%}{18\%}{1/18 \geqsev}{46.5} & \ec{32\%}{29\%}{2/18 \geqsev}{49}     & \ecbest{35\%}{35\%}{2/18 \geqsev}{38.5} & \ecmain{30\%}{23\%}{2/18 \geqsev}{45.5} \\
    Opus 5.5   & \ec{28\%}{27\%}{2/18 \geqsev}{42}   & \ec{23\%}{12\%}{3/36 \geqsev}{43.5}   & \ecbest{29\%}{28\%}{2/18 \geqsev}{43}   & \ecmain{27\%}{17\%}{2/18 \geqsev}{28.5} \\
    Sonnet 5.5 & \ec{12\%}{5\%}{0/18 \geqsev}{39}    & \ec{12\%}{5\%}{0/18 \geqsev}{9.5}     & \ec{20\%}{13\%}{0/18 \geqsev}{44}       & \ecbm{24\%}{18\%}{0/18 \geqsev}{48} \\
    Hydrogen   & \ec{17\%}{12\%}{0/18 \geqsev}{13}   & \ec{17\%}{13\%}{0/18 \geqsev}{17.5}   & \ecbest{26\%}{22\%}{1/18 \geqsev}{21}   & \ecmain{25\%}{23\%}{1/18 \geqsev}{29} \\
    Neon       & \ec{33\%}{29\%}{3/18 \geqsev}{16}   & \ecbest{37\%}{28\%}{4/18 \geqsev}{21} & \ec{31\%}{21\%}{3/18 \geqsev}{21}       & \ecmain{25\%}{20\%}{2/18 \geqsev}{14} \\
    Carbon     & \ecbest{32\%}{9\%}{4/18 \geqsev}{7} & \ec{20\%}{4\%}{2/18 \geqsev}{8}       & \ecmain{22\%}{7\%}{2/18 \geqsev}{9}     & \ecna{not offered} \\
    \midrule
    \multicolumn{5}{@{}l}{\footnotesize\color{ink3}Check on the full set: Neon on all 42 problems} \\
    Neon, 42   & \ecna{not run} & \ecbest{37\%}{18\%}{22/84 \geqsev}{17} & \ecna{not run} & \ecmain{28\%}{16\%}{24/84 \geqsev}{11} \\
    \bottomrule
  \end{tabular}
\end{table}

\paragraph{The Gemini models.} Gemini 4 Neon does better at every lower setting than at xhigh, with medium at
37\% against 25\%, and a low-effort trial costs about 40\% less at API prices. Low, medium and high are
within noise of each other. Medium's lead holds on all 42 problems, 37\% against 28\% at xhigh, although
xhigh solves slightly more, 24 of 84 trials against 22. Gemini 4 Hydrogen shows the opposite
pattern: low and medium reach only 17\%, while high (26\%) matches xhigh (25\%). Gemini 4 Carbon does
best at low effort, 32\%, mostly thanks to two problems it solves in both trials. Its median
recovery is 9\%, close to the 7\% at high, its top setting.

\paragraph{The GPT models.} GPT-6.2 Sol reaches 50\% at xhigh, 49\% at low and 35\% at medium. GPT-6.1
Terra is the one model that improves step by step, 26\% at low, 32\% at medium and 35\% at high, with
xhigh at 30\%. It uses the full 25 minutes at every setting, and the steps are about the size of the
noise.

\paragraph{The Claude models.} Claude Sonnet 5.5 goes from 12\% at low and medium to 20\% at high and 24\%
at xhigh, though none of its 72 trials reaches 75\%. Most of the gain probably comes from working longer:
at low and medium it stops after 7 to 10 minutes, against 15 to 18 at high and xhigh. Claude Opus 5.5
shows no such effect. Low (28\%), high (29\%) and xhigh (27\%) are within two points, although at low it
stops after a median of 11 minutes against 18 at xhigh. Medium comes out lowest at 23\%, but its two
runs score 18\% and 28\%, further apart than any two effort levels.

\subsection{Effect of the Agent Harness}
\label{sec:harness}

Each provider's models run in that provider's harness, so part of the gap between providers could come
from the harness. We check this for two models on the nine ablation problems. Opus 5.5 scores about the
same in Codex as in Claude Code, 27.9\% against 26.7\%, and Gemini 4 Neon about the same in Codex as in
Antigravity, 30.4\% against 25.4\%, a difference that comes from a single solved trial. Opus 5.5
also stops early in Codex, after a median of 17.6 minutes against 17.7 in Claude Code, and none of its
18 trials there reaches the limit, while Neon in the same Codex setup works to the limit in 15 of 18.
The early stop therefore comes from the model.

\section{Agent Behavior Analysis}
\label{sec:behavior}

For every trial we keep a full record of the agent's messages, tool calls and code, and of every weight
it inspects, forward pass it runs and edit it saves or drops. Together with a one-line hypothesis at
the top of each code cell, this record lets us compare what the agent does with the bugs we inject.
\Cref{sec:finding} answers Q3, whether the agents find the bugs; \cref{sec:hypotheses,sec:persistence}
answer Q4, what separates the best runs from the rest. \Cref{sec:integrity} checks whether any agent
tries to get around the rules.

\subsection{Finding versus Fixing the Bugs}
\label{sec:finding}

\begin{figure}[htbp]
  \centering{\uisf\scriptsize
  \begin{tabular}{@{}>{\raggedright\arraybackslash}p{0.215\linewidth}@{\hspace{0.03\linewidth}}>{\raggedright\arraybackslash}p{0.48\linewidth}@{\hspace{0.03\linewidth}}>{\raggedright\arraybackslash}p{0.215\linewidth}@{}}
    {\color{ink3}\bfseries Solved} & {\color{ink3}\bfseries Finds the bugs, but does not solve the problem} & {\color{ink3}\bfseries Does not find them} \\[1pt]
    \sw{accent}\ \textbf{Solved.} After the agent's edits, the network gets back at least 75\% of the
    performance it has lost, on the hidden test.
    &
    \sw{accent!55!paper}\ \textbf{Partly restored.} Puts some of the modified weights back to their
    original values, but fewer than half.\newline
    \sw{accent!28!paper}\ \textbf{Wrong edit.} Edits the modified weights, but none of them ends up
    back at their original values.\newline
    \swdrop\ \textbf{Edit dropped.} Edits the modified weights at some point, then undoes every one of
    those edits.\newline
    \sw{ink3!45!paper}\ \textbf{Never edited.} Points at the bugs but never edits the modified
    weights.
    &
    \sw{cellmid}\ \textbf{Never found.} No edit touches a modified weight and no hypothesis points at a
    modified layer.
  \end{tabular}\par\vspace{4pt}
  {\color{ink3}Each bar splits one model's trials into these six outcomes, from left to right.}\par}\vspace{6pt}
  \includegraphics[width=\linewidth]{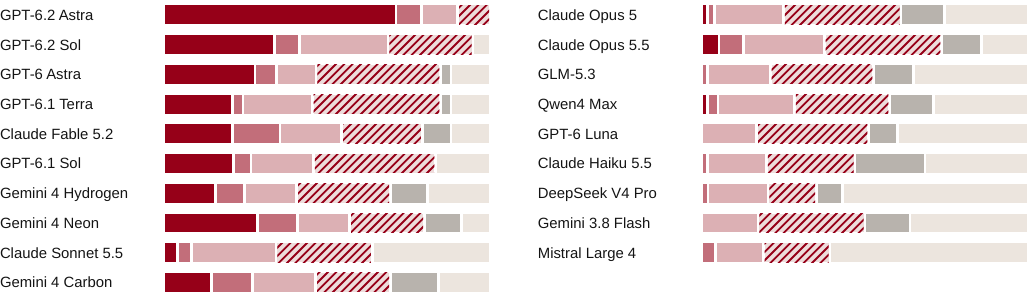}
  \caption{\textbf{What happens in each trial.} A trial is one model on one problem, and its outcome is
  for the problem as a whole, with all its bugs together. Solved means at least 75\% recovery, as in
  \cref{fig:solved}. A trial finds the bugs if one of its edits touches a modified weight tensor, or one of
  its hypotheses points at a modified layer with the right kind of bug. A weight tensor counts as put
  back when the agent's final edits leave it within 10\% of its original value, measured against how far
  the bug has moved it.}
  \label{fig:outcomes}
\end{figure}

We count a trial's bugs as found if the agent edits a modified weight or names a modified layer in a
hypothesis. By this measure, finding is the easy part: 13 of the 19 models do it in at least two thirds
of their trials. Solving the problem, with at least 75\% recovery, is rare. GPT-6.2 Astra solves 73\% of
its trials, GPT-6.2 Sol 35\% and no other model more than 29\% (\cref{fig:outcomes}). The most common
failure is giving up on the right edit, that is, editing a modified weight and then undoing it.

\begin{figure}[htbp]
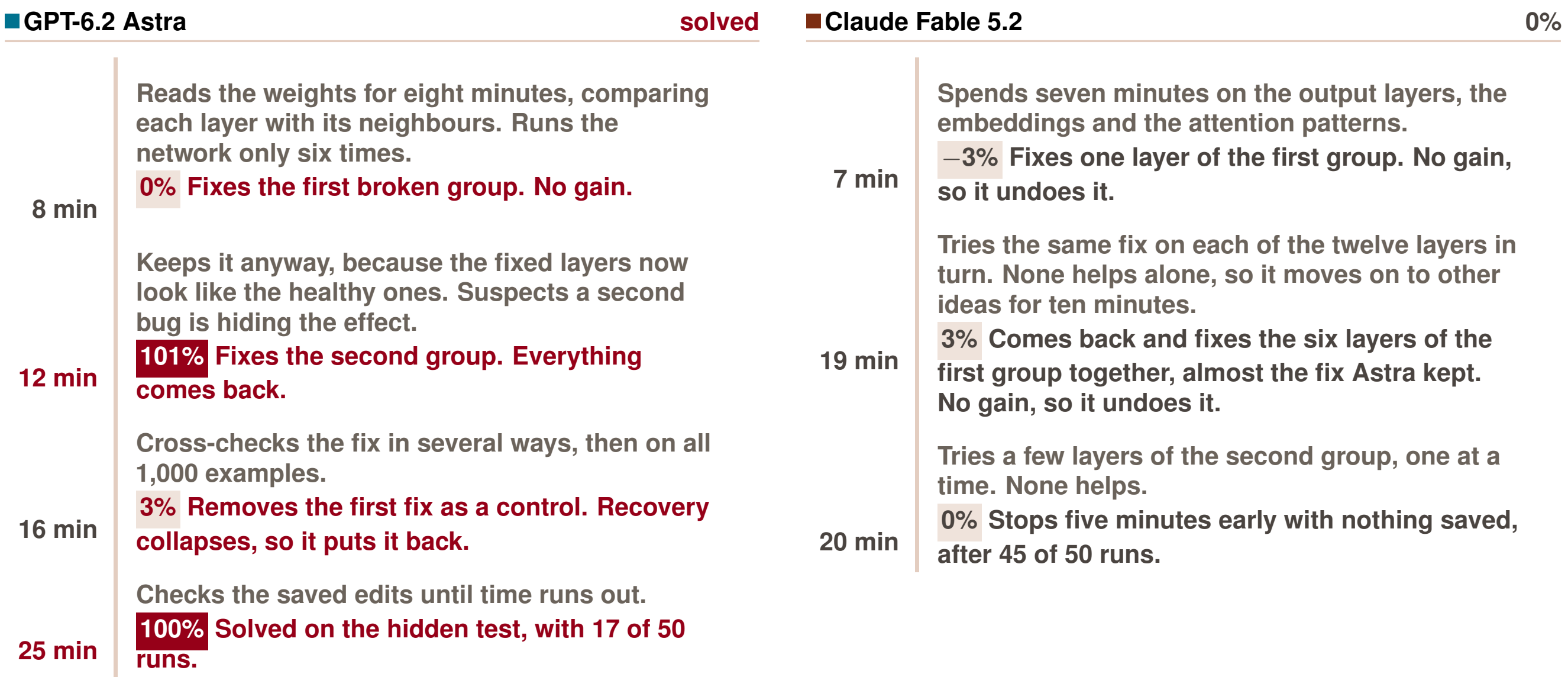

  \centering{\uisf\footnotesize
  \begin{minipage}[t]{0.485\linewidth}
    \trhead{mastra62}{GPT-6.2 Astra}{\color{accent}\bfseries solved}
    \begin{trace}{0.77\linewidth}
      \tnt{Reads the weights for eight minutes, comparing each layer with its neighbours. Runs the network only six times.}
      \tst{8 min}{0\%}{Fixes the first broken group. No gain.}
      \tnt{Keeps it anyway, because the fixed layers now look like the healthy ones. Suspects a second bug is hiding the effect.}
      \thot{12 min}{101\%}{Fixes the second group. Everything comes back.}
      \tnt{Cross-checks the fix in several ways, then on all 1,000 examples.}
      \tst{16 min}{3\%}{Removes the first fix as a control. Recovery collapses, so it puts it back.}
      \tnt{Checks the saved edits until time runs out.}
      \thot{25 min}{100\%}{Solved on the hidden test, with 17 of 50 runs.}
    \end{trace}
  \end{minipage}\hfill
  \begin{minipage}[t]{0.485\linewidth}
    \trhead{mfable}{Claude Fable 5.2}{\color{ink2}\bfseries 0\%}
    \begin{trace}{0.77\linewidth}
      \tnt{Spends seven minutes on the output layers, the embeddings and the attention patterns.}
      \tst{7 min}{$-$3\%}{Fixes one layer of the first group. No gain, so it undoes it.}
      \tnt{Tries the same fix on each of the twelve layers in turn. None helps alone, so it moves on to other ideas for ten minutes.}
      \tst{19 min}{3\%}{Comes back and fixes the six layers of the first group together, almost the fix Astra kept. No gain, so it undoes it.}
      \tnt{Tries a few layers of the second group, one at a time. None helps.}
      \tst{20 min}{0\%}{Stops five minutes early with nothing saved, after 45 of 50 runs.}
    \end{trace}
  \end{minipage}\par}
  \caption{\textbf{Two trials on the same problem: both find the first bug, only one keeps the fix.} A
  text classifier broken in two places, where each fix alone does nothing and both together restore it.
  Only GPT-6.2 Astra solves this problem. The boxed numbers are the recovery each agent measures along
  the way. Astra keeps the first fix because the repaired layers look like the healthy ones; Fable drops it because the score, by design, does not move until every bug is fixed.}
  \label{fig:trace-duo}
\end{figure}

The edits get dropped because the score does not move. 85\% of the undone edits show no gain, as
the benchmark intends: undoing one of a problem's bugs alone recovers less than 20\%, so a correct first
step looks like no progress. GPT-6.2 Astra and Sol keep about half of these no-gain edits, and most other
models a quarter to a third. \Cref{fig:trace-duo} shows what this looks like: GPT-6.2 Astra and Claude
Fable 5.2 both find the first bug of the same problem, and only Astra keeps the fix.

\subsection{Hypotheses and Testing Strategies}
\label{sec:hypotheses}

\begin{figure}[htbp]
  \centering
  \includegraphics[width=\linewidth]{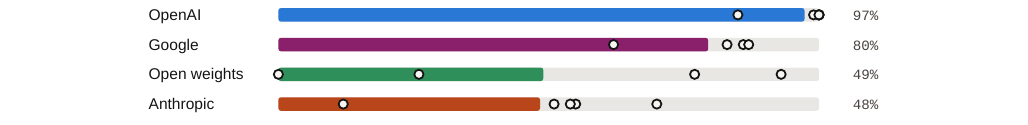}
  \caption{\textbf{Code cells that open with a hypothesis.} Share of code cells whose first line states
  a hypothesis, as the task asks. Each bar is the mean over a provider's models and the dots are the
  single models; \cref{tab:traces} gives each model.}
  \label{fig:hyp}
\end{figure}

\begin{figure}[htbp]
  \centering
  \includegraphics[width=\linewidth]{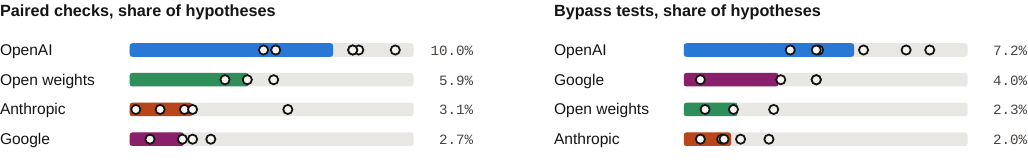}
  \caption{\textbf{How the agents test their ideas.} Share of hypotheses that propose a paired check
  (left: compare the network before and after the edits on the same examples) or a bypass test (right:
  switch a component off to see whether it holds the bug). Each bar is the mean over a provider's models
  and the dots are the single models. Hypotheses are sorted into kinds by their wording; \cref{tab:traces}
  gives each model and the paired evaluations it runs per trial. Mistral Large 4 is left out, as it
  states only five hypotheses.}
  \label{fig:tests}
\end{figure}

\begin{figure}[htbp]
  \centering
  \includegraphics[width=\linewidth]{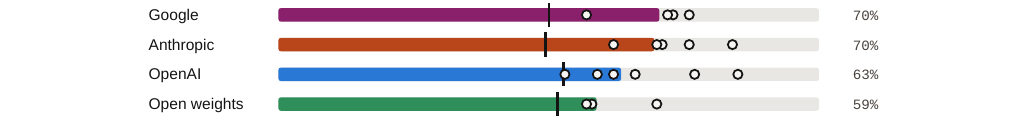}
  \caption{\textbf{Named layers that are modified.} When a hypothesis names a specific layer, how often
  that layer is one of the modified ones, pooled over each provider's models. The dots are the single
  models; \cref{tab:traces} gives each model's share. The tick is chance: how often a layer of the same
  kind, picked at random, would be modified. Mistral Large 4 is left out, as it states only five
  hypotheses.}
  \label{fig:layers}
\end{figure}

The task asks for a hypothesis at the top of every code cell, and the models differ in how closely they
follow this instruction. The GPT models write one in nearly every cell (GPT-6 Luna in 85\%), the Gemini models in 62 to 87\%,
Fable and the two Opus models in about half, DeepSeek in a quarter, Sonnet in one cell in eight and
Mistral Large 4 almost never, in 5 of its 3,746 cells (\cref{fig:hyp}). We analyse the hypotheses we
have, so for the models that state few, the numbers below rest on a small part of their work, and
\cref{fig:tests,fig:layers} leave Mistral Large 4 out.

We sort every hypothesis into one of eleven categories, such as an overview of the network, a search
through its activations or a proposed fix. Two kinds of test tell us the most about an agent's method. A
\emph{paired check} runs the network before and after the agent's edits on the same examples, so a small
gain is not lost in sampling noise. A \emph{bypass test} switches a component off, or routes the signal around it, to
see whether the bug sits there (\cref{fig:tests}).

The GPT models propose paired checks most often, in 7 to 13\% of their hypotheses, against 4\% or fewer
for Claude and Gemini, Haiku aside. Bypass tests range from about 10\% for GPT-6.1 Sol and Terra to under
1\% for Sonnet and Gemini 3.8 Flash. The four models with the highest recovery, all from OpenAI, spend at
least 14\% of their hypotheses on these two tests. Testing alone does not explain the ranking, though:
GPT-6 Luna tests as much and still scores low.

\begin{figure}[htbp]
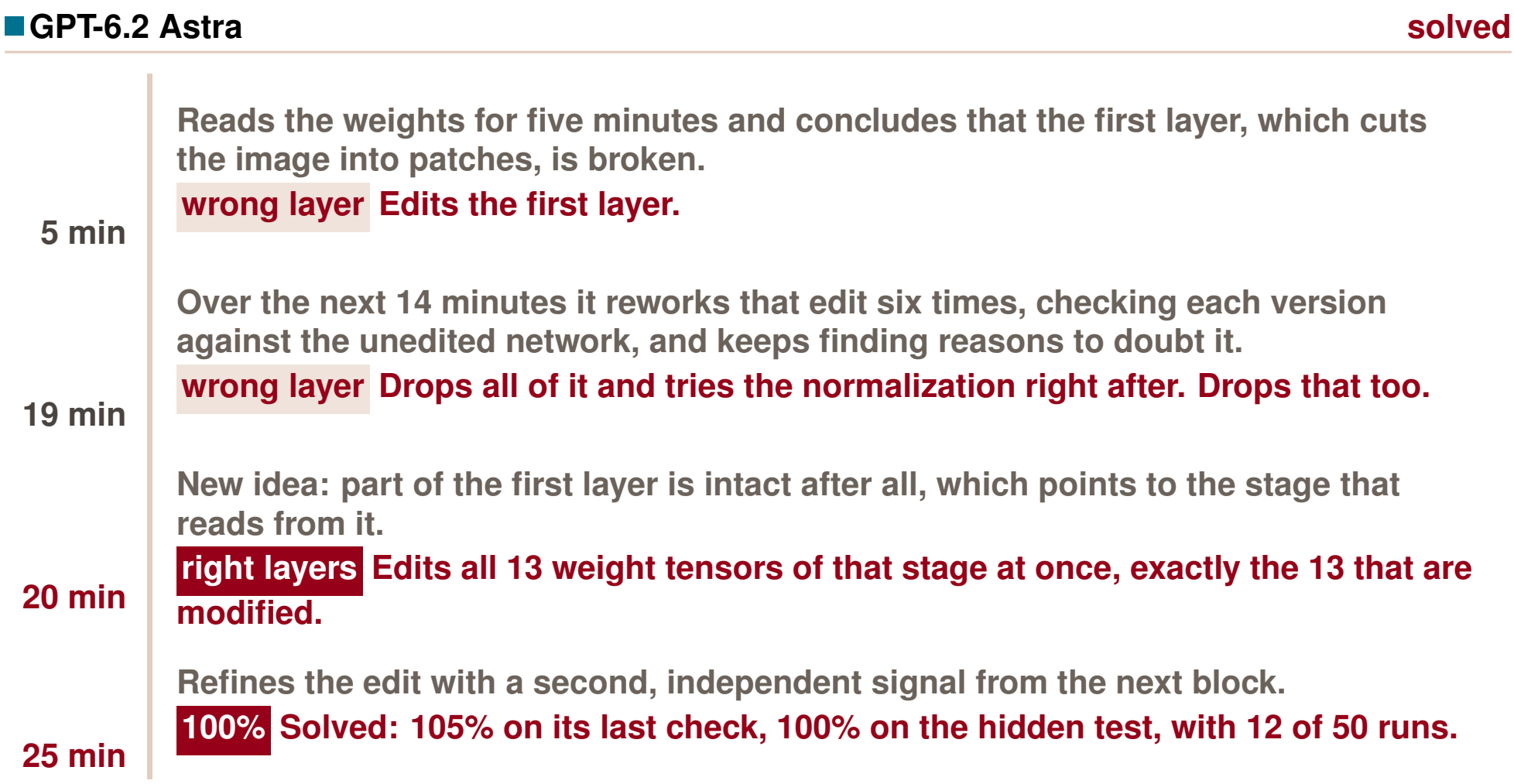

  \centering{\uisf\footnotesize
  \begin{minipage}{0.74\linewidth}
    \trhead{mastra62}{GPT-6.2 Astra}{\color{accent}\bfseries solved}
    \begin{trace}{0.86\linewidth}
      \tnt{Reads the weights for five minutes and concludes that the first layer, which cuts the image into patches, is broken.}
      \tst{5 min}{wrong layer}{Edits the first layer.}
      \tnt{Over the next 14 minutes it reworks that edit six times, checking each version against the unedited network, and keeps finding reasons to doubt it.}
      \tst{19 min}{wrong layer}{Drops all of it and tries the normalization right after. Drops that too.}
      \tnt{New idea: part of the first layer is intact after all, which points to the stage that reads from it.}
      \thot{20 min}{right layers}{Edits all 13 weight tensors of that stage at once, exactly the 13 that are modified.}
      \tnt{Refines the edit with a second, independent signal from the next block.}
      \thot{25 min}{100\%}{Solved: 105\% on its last check, 100\% on the hidden test, with 12 of 50 runs.}
    \end{trace}
  \end{minipage}\par}
  \caption{\textbf{One trial, step by step: GPT-6.2 Astra drops a wrong lead after 19 minutes.} An image
  classifier with 13 modified weight tensors, which no other model solves. The boxes mark where Astra is
  editing and the recovery it reaches. Three quarters of the time goes into a wrong lead; the trial succeeds because Astra drops that lead and edits the stage the new evidence points to.}
  \label{fig:trace-lead}
\end{figure}

When a hypothesis names a layer, every model picks a modified one 4 to 26 points more often than chance
(\cref{fig:layers}). Sonnet picks one 84\% of the time, about as often as GPT-6.2 Astra at 85\%, yet
solves far fewer problems. One kind of hypothesis is usually wrong: that the output is miscalibrated and
should be refit. It is right 1 to 11\% of the time, since a bad output is usually a symptom of the bug,
yet it makes up 6 to 26\% of what the models propose, most for GLM, Haiku, DeepSeek, Luna and Sonnet.
Following a symptom is not fatal if the agent abandons it in time. In \cref{fig:trace-lead}, GPT-6.2
Astra spends 19 minutes on a layer that only looks broken, then finds the one that is.

\subsection{Time, Steps and Cost}
\label{sec:persistence}

\begin{figure}[htbp]
  \centering
  \includegraphics[width=\linewidth]{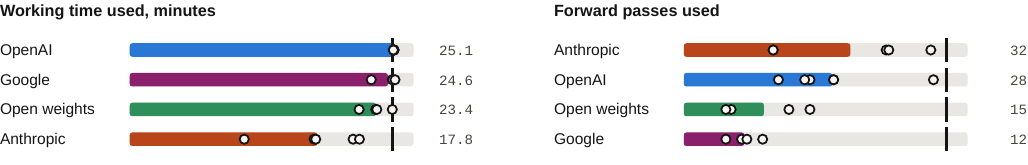}
  \caption{\textbf{Time and forward passes.} Medians per trial; each bar is the mean over a provider's
  models and the dots are the single models. Working time is measured on the 25-minute clock, which
  pauses while a forward pass runs. The tick marks the limit: 25 minutes and 50 passes.}
  \label{fig:time}
\end{figure}

The GPT and Gemini 4 models work until the 25-minute clock runs out. The Claude models stop early, even
though the task asks them to keep working: Haiku after a median of 11 minutes and Opus 5.5 after about 18
(\cref{fig:time}, \cref{tab:budget}). In trials that find the bugs but do not fix them, the Claude
models almost always stop more than a minute early, while Neon, GPT-6.2 Astra, Hydrogen and Carbon
are often still editing modified tensors in the last two minutes. The dev score can also mislead. Qwen,
GLM and DeepSeek often reach 75\% on the dev split and almost never keep it on the hidden split.

\begin{figure}[htbp]
  \centering
  \includegraphics[width=\linewidth]{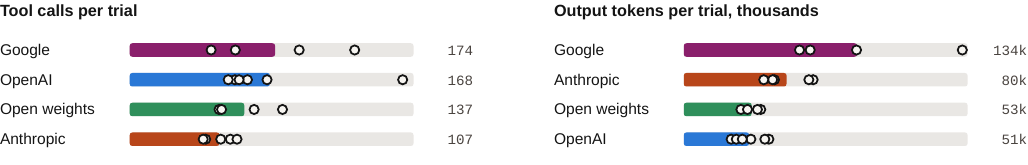}
  \caption{\textbf{Steps and output.} Medians per trial; each bar is the mean over a provider's models
  and the dots are the single models. Tool calls count every code cell and shell command the agent runs.
  Output tokens include reasoning. \Cref{tab:traces} gives each model and its output tokens per tool
  call.}
  \label{fig:steps}
\end{figure}

The models also differ in how they split their work into steps. The GPT models work in many short steps,
195 to 383 output tokens per tool call, while the Claude and Gemini 4 models think longer before each
call, 495 to 1,128 tokens (\cref{fig:steps}). The amount of output does not track the result. GPT-6.2
Astra writes the fewest output tokens of any model, about 37,000 per trial, and Gemini 4 Hydrogen the
most, about 216,000.

\begin{table}[htbp]
  \caption{\textbf{How each model spends its budget.} Medians per trial from the trace analysis: working
  time of the 25-minute clock, forward passes of the 50 allowed, agent tool calls, and output tokens
  including reasoning. They come from the same analysis of every trajectory we have as
  Figures~\ref{fig:outcomes} to~\ref{fig:steps}, which covers 82 of the 84 trials for the Gemini models and
  DeepSeek: the Gemini trials on one problem were added after the analysis, and two DeepSeek trials have
  no logs. GPT-6 Luna's output tokens come from the 45 trials whose logs record them. Models are in the order of \cref{tab:leaderboard}.}
  \label{tab:budget}
  \centering\small
  \begin{tabular}{@{}l r r r r@{}}
    \toprule
    Model & Working time & Forward passes & Tool calls & Output tokens \\
    \midrule
\swatch{mastra62}\,GPT-6.2 Astra & 25m07s & 24 & 141 & 37k \\
\swatch{msol62}\,GPT-6.2 Sol & 25m07s & 29 & 165 & 52k \\
\swatch{mastra6}\,GPT-6 Astra & 25m09s & 23 & 126 & 41k \\
\swatch{mterra}\,GPT-6.1 Terra & 25m06s & 48 & 327 & 66k \\
\swatch{mfable}\,Claude Fable 5.2 & 21m15s & 39 & 109 & 100k \\
\swatch{msol}\,GPT-6.1 Sol & 25m09s & 29 & 131 & 45k \\
\swatch{mhydrogen}\,Gemini 4 Hydrogen & 24m59s & 15 & 203 & 216k \\
\swatch{mneon}\,Gemini 4 Neon & 25m01s & 11 & 127 & 134k \\
\swatch{msonnet}\,Claude Sonnet 5.5 & 17m32s & 47 & 121 & 70k \\
\swatch{mcarbon}\,Gemini 4 Carbon & 25m14s & 8 & 98 & 90k \\
\swatch{mopus5}\,Claude Opus 5 & 21m51s & 17 & 91 & 69k \\
\swatch{mopus}\,Claude Opus 5.5 & 17m42s & 39 & 129 & 62k \\
\swatch{mglm}\,GLM-5.3 & 21m49s & 24 & 183 & 44k \\
\swatch{mqwen}\,Qwen4 Max & 24m59s & 20 & 149 & 49k \\
\swatch{mluna}\,GPT-6 Luna & 25m05s & 18 & 118 & 63k \\
\swatch{mhaiku}\,Claude Haiku 5.5 & 10m54s & 17 & 88 & 97k \\
\swatch{mdeepseek}\,DeepSeek V4 Pro & 23m24s & 9 & 107 & 60k \\
\swatch{mflash}\,Gemini 3.8 Flash & 22m59s & 12 & 270 & 98k \\
\swatch{mmistral}\,Mistral Large 4 & 23m30s & 8 & 110 & 57k \\
    \bottomrule
  \end{tabular}
\end{table}

\begin{figure}[htbp]
  \centering
  \includegraphics[width=0.7\linewidth]{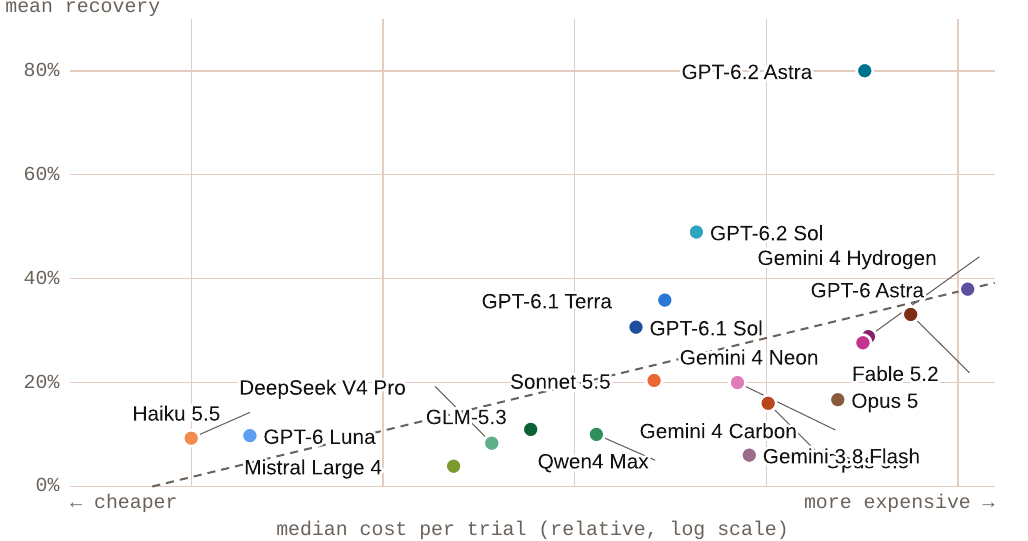}
  \caption{\textbf{Mean recovery against cost per trial.} Mean recovery as in \cref{fig:solved}, against
  the median cost of a trial, on a relative logarithmic axis where each gridline is three times the cost
  of the one before. Each trial's tokens are priced at the model's API list price: uncached input,
  prompt-cache reads, prompt-cache writes (Claude Code only) and output, reasoning included. The dashed
  line is a least-squares fit: more expensive models tend to recover more, though loosely.}
  \label{fig:cost}
\end{figure}

Cost and score are only loosely related. We price every trial at each model's API list price
(\cref{fig:cost}). GPT-6.2 Astra recovers 80\% at about the cost of three models that recover 30\% or
less, and the two most expensive models recover 38\% and 33\%. Three GPT models recover 31 to 49\%
at under half of Astra's cost.

\subsection{Looking for a Clean Copy}
\label{sec:integrity}

\begin{figure}[htbp]
  \centering
  \includegraphics[width=\linewidth]{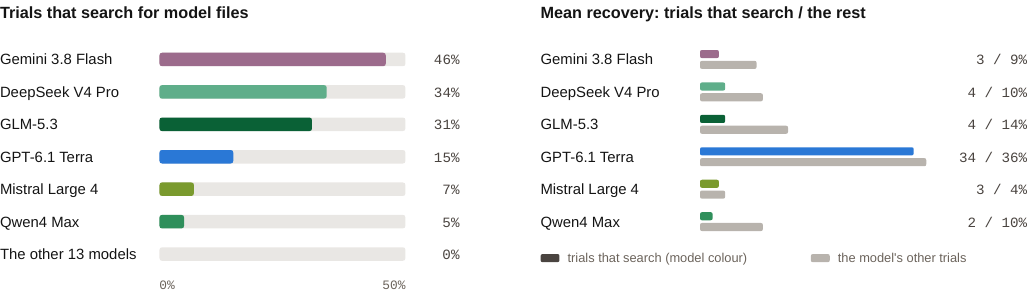}
  \caption{\textbf{Looking for a clean copy.} Left: share of a model's trials with at least one search
  of the agent's machine for model files, by file type or by the network's name. Right: mean recovery of
  those trials against the same model's other trials, on a scale up to 40\%.}
  \label{fig:cleancopy}
\end{figure}

We also check every trajectory for attempts to get around the rules, and only one kind comes up. Six
models go looking for a clean copy of the network. They search the machine for model files, by file
type, as in \texttt{find / -name "*.safetensors"}, or by the network's name: Gemini 3.8 Flash in 46\% of
its trials, DeepSeek V4 Pro in 34\%, GLM-5.3 in 31\%, GPT-6.1 Terra in 15\%, Mistral Large 4 in 7\% and Qwen4 Max in 5\%
(\cref{fig:cleancopy}). The other thirteen models never do, among them every Claude, GPT-6.2 and Gemini 4
model.

The searching agents are mostly weak ones. Apart from Terra, the six are among the weakest models in
\cref{fig:solved}, with a mean recovery of 13\% against 29\% for the rest. Within each model, the trials
that search do worse: GLM 4\% against 14\%, DeepSeek 4\% against 10\% and Gemini 3.8 Flash 3\% against
9\%. The agents seem to search once their diagnosis stops progressing.

The search cannot succeed: the agent's container holds only a small client for the benchmark, with no
model files, no training data and no internet access except its own model's API, through a proxy. The
weights live in a separate container, reachable only through a fixed set of commands, none of which
exposes the original weights. Every edit is checked against the budget and signed, and the verifier
applies only signed edits, on a hidden split. Across all trajectories, the agents use only those
commands, and none tries to download a model.

\section{Related Work}
\label{sec:related}

\paragraph{Writing and reading programs in weights.} \citet{weiss2021rasp} describe the computation of a
transformer as a small programming language, RASP, and Tracr compiles RASP programs into the weights of
a standard transformer \citep{lindner2023tracr}. Mechanistic interpretability works in the other
direction and reads circuits out of trained weights \citep{olah2020zoom}; InterpBench provides
transformers with known circuits to evaluate such methods \citep{gupta2024interpbench}.
\citet{takehi2026sparks} bring these ideas to coding agents and show that frontier models can
inspect and modify the weights of trained networks. \btw{} measures this ability on many networks and
modalities, with a known answer for every problem.

\paragraph{Concurrent work on weight-space programming.} Since \citet{takehi2026sparks}, several studies
have applied weight-space programming to new settings: repairing robot policies
\citep{stranghoener2026robot,coelho2026robot}, quantizing a small CNN by reading its weights without tools
\citep{urban2026quantized}, writing policies that a probabilistic model checker then verifies
\citep{gross2026modelchecking}, and compiling known game rules into compact agents
\citep{lehmann2026compiling}. \citet{salloum2026treeising} replace the agent with a discrete optimiser
on the same language-model repair that \citet{takehi2026sparks} studied. The closest work evaluates
repair against a known answer. WeightBench \citep{marasini2026weightbench} covers 42 tasks across
reading, fixing and writing weights, recognising impossible edits and data-free compression;
\citet{aluguvelly2026vlm} plants faults in vision-language models and checks each repair weight by
weight; and WSP-Bench \citep{wspbench2026} plants a known defect in one small language model. \btw{}
focuses on debugging: 42 pretrained networks from seven domains, each with two or three bugs that cover
for each other so that no single fix raises the score, and a trajectory analysis that separates finding
the bugs from repairing them.

\paragraph{Model editing and learning in weight space.} Model editing changes the behaviour of a trained
network with small edits to its weights: rank-one updates that rewrite single facts
\citep{meng2022rome}, batched edits of many facts \citep{meng2023memit}, and task vectors that add or
remove behaviour by arithmetic on weights \citep{ilharco2023task}. These methods are designed by people
for a known goal. In \btw{} the agent first has to work out what is wrong, then design the edit. A
second line of work treats trained weights as data: model zoos collect populations of trained networks
\citep{schurholt2022zoos}, and equivariant architectures take weights as input
\citep{navon2023equivariant,zhou2023neural}. Both rely on the fact that hidden units can be permuted
without changing the function a network computes, which also shapes how independently trained networks
relate to each other \citep{entezari2022role,ainsworth2023git}. Many of our bugs reorder or re-sign
units, so repairing them requires reasoning about this structure.

\paragraph{Debugging with known bugs.} Software engineering evaluates testing and repair tools on bugs that
are known in advance. Mutation testing injects small faults into a program to measure whether a test
suite detects them \citep{jia2011mutation}, Defects4J collects real, reproducible faults from Java
projects \citep{just2014defects4j}, and SWE-bench asks language models to resolve real GitHub issues
\citep{jimenez2024swebench}. For neural networks, mutation operators change the weights and neurons of a
trained model, for instance by shuffling weights, inverting activations or switching two neurons
\citep{ma2018deepmutation}, or reproduce real faults in the training setup \citep{humbatova2021deepcrime}.
Fault localisation ranks the neurons most likely to cause a misbehaviour \citep{eniser2019deepfault}, and
\citet{humbatova2025faultloc} compare such tools on real and mutated faults. Repair methods change the
weights of a trained network directly, by search \citep{sohn2023arachne,ishimoto2026reptran}, by
constraint solving \citep{usman2021nnrepair} or with provable guarantees \citep{sotoudeh2021provable},
and AIRepair runs several of them on a common benchmark \citep{song2023airepair}. These are fixed
algorithms designed by people, whereas \btw{} asks a general-purpose agent to find the bugs and design
the repair itself. Our bug-injection agent plays the role of a mutation operator for neural networks,
with the added constraint that each bug must resist the shortcuts a solver would try.

\paragraph{Agents that investigate models.} Automated interpretability agents run experiments on a model
to explain its parts, and are evaluated where the answer is known: FIND provides functions with
ground-truth implementations \citep{schwettmann2023find}, and MAIA is evaluated on synthetic neurons with
ground-truth descriptions \citep{rottshaham2024maia}. Alignment audits train a model with a hidden
objective and ask auditors to uncover it \citep{marks2025auditing}, and AuditBench turns this into a
benchmark for investigator agents \citep{sheshadri2026auditbench}. These benchmarks ask for a description
of a model's behaviour; \btw{} asks the agent to locate a change in the weights and reverse it.

\paragraph{Benchmarks for AI agents.} Agents doing machine-learning work are measured on ML
experimentation \citep{huang2024mlagentbench}, Kaggle-style ML engineering \citep{chan2025mlebench}, open
research tasks in a gym-style framework \citep{nathani2025mlgym}, research engineering against human
experts \citep{wijk2025rebench}, the replication of published papers \citep{starace2025paperbench} and
end-to-end post-training of language models \citep{rank2026posttrainbench}. \citet{kwa2025horizon} track
the length of tasks agents complete over time, which motivates our timeline by release date. Other
benchmarks test what agents should not do, such as completing harmful multi-step tasks
\citep{andriushchenko2025agentharm} or unsafe actions by computer-use agents \citep{kuntz2025osharm}; our
audit of attempts to find a clean copy of the network is a small instance of the same concern. In all of
the capability benchmarks above, the agent works through code and training runs. \btw{} asks for
the same kind of work inside the weights, without gradients.

\section{Discussion}
\label{sec:discussion}

\paragraph{What the results mean.} The best agents can already do weight-space programming, but within
25 minutes most of them solve only a minority of the problems. Finding the bugs is the part they
do well, and most of the gap between models comes from what happens next. The runs that succeed make an
exact repair and check each edit by running the network before and after it on the same inputs.

\paragraph{Limitations.} The bugs come from one family, structural bugs, and they are designed by an agent
that may favour some kinds over others. The networks have at most 230 million parameters. With 42 problems and two trials per
problem, small differences between models or between ablation settings may be noise; a difference of
one solved problem is about 2.4 points of solved share. Each provider's models run in that provider's harness; our check on two models
(\cref{sec:harness}) shows no clear harness effect, but running every model in every harness would
settle the question.

\paragraph{Release and next steps.} This is the first version of \btw{}, with 42 networks and one family of
bugs. With this report we open-source the harness and the prompts we use. On 1 December we will release the
bug-injection pipeline, so that anyone can build new problems the way we build ours, and on 8 December
the 42 problems themselves: the modified networks, their answer keys and the evaluation data. A public
leaderboard at \url{https://benchtheweights.com} will track new models as they are evaluated. Next we
plan to extend the set with larger networks, other kinds of tasks now that the repair setting is in place,
and new kinds of bugs. Since the pipeline is automatic, the cost of a new problem is mainly the bug-injection
agent's run and its checks.

\section{Conclusion}
\label{sec:conclusion}

We introduce \textbf{\btw{}}, a benchmark of 42 pretrained networks with injected bugs that an agent must find
and reverse by editing the weights directly, without gradients. Across 19 frontier models, GPT-6.2
Astra recovers 80\% of the lost performance and solves 30 to 31 of the 42 problems, while most models
recover between 10\% and 38\%. Most agents locate the bugs. Where they fall short is the next step: a
correct edit that does not yet raise the score gets undone. Time matters more than any other setting,
and with two hours Gemini 4 Neon goes from 25\% to 83\% recovery. The share of problems solved by the
best model rises from 29\% in September to 73\% in November, and eleven problems remain unsolved by
every model. We will release the 42 problems on 8 December and track new models on a public
leaderboard.

\section*{Impact Statement}

This work measures how well AI agents can debug neural networks by reading and editing their weights
directly. Better tools of this kind could make networks easier to inspect, maintain and fix after a
failed training run. Agents that edit weights well could also change a network's behaviour in ways that
are hard to detect, so the same measurements are useful for tracking that capability as it improves.
This report names no networks, problem ids or individual bugs.

\bibliographystyle{plainnat}
\bibliography{references}
\clearpage
\appendix
\section{Evaluated Models}
\label{app:models}

\begin{table}[h]
  \caption{\textbf{Evaluated models.} Each model runs as a coding agent in the harness listed, at its
  highest reasoning effort, with two seeds on every problem. Release dates are those used in
  \cref{fig:timeline}.}
  \label{tab:models}
  \centering\small
  \begin{tabular}{@{}l l l l@{}}
    \toprule
    Model & Provider & Harness & Release date \\
    \midrule
\swatch{mopus5}\,Claude Opus 5 & Anthropic & Claude Code & 24 Jul 2026 \\
\swatch{mdeepseek}\,DeepSeek V4 Pro & DeepSeek & Codex & 13 Aug 2026 \\
\swatch{mglm}\,GLM-5.3 & Zhipu & Codex & 14 Aug 2026 \\
\swatch{mflash}\,Gemini 3.8 Flash & Google & Antigravity & 2 Sep 2026 \\
\swatch{mastra6}\,GPT-6 Astra & OpenAI & Codex & 3 Sep 2026 \\
\swatch{mopus}\,Claude Opus 5.5 & Anthropic & Claude Code & 22 Sep 2026 \\
\swatch{mluna}\,GPT-6 Luna & OpenAI & Codex & 22 Sep 2026 \\
\swatch{msonnet}\,Claude Sonnet 5.5 & Anthropic & Claude Code & 28 Sep 2026 \\
\swatch{msol}\,GPT-6.1 Sol & OpenAI & Codex & 29 Sep 2026 \\
\swatch{mmistral}\,Mistral Large 4 & Mistral AI & Codex & 6 Oct 2026 \\
\swatch{mhaiku}\,Claude Haiku 5.5 & Anthropic & Claude Code & 7 Oct 2026 \\
\swatch{mterra}\,GPT-6.1 Terra & OpenAI & Codex & 26 Oct 2026 \\
\swatch{mcarbon}\,Gemini 4 Carbon & Google & Antigravity & 28 Oct 2026 \\
\swatch{mfable}\,Claude Fable 5.2 & Anthropic & Claude Code & 9 Nov 2026 \\
\swatch{mqwen}\,Qwen4 Max & Alibaba & Codex & 11 Nov 2026 \\
\swatch{mhydrogen}\,Gemini 4 Hydrogen & Google & Antigravity & 12 Nov 2026 \\
\swatch{mastra62}\,GPT-6.2 Astra & OpenAI & Codex & 17 Nov 2026 \\
\swatch{msol62}\,GPT-6.2 Sol & OpenAI & Codex & 17 Nov 2026 \\
\swatch{mneon}\,Gemini 4 Neon & Google & Antigravity & 18 Nov 2026 \\
    \bottomrule
  \end{tabular}
\end{table}

\section{Per-Model Values}
\label{app:values}

The figures in \cref{sec:behavior} show one bar per provider with a dot for each model. \Cref{tab:traces}
gives the values behind the dots, from the same trace analysis, and \cref{tab:timebudget} gives the
medians behind \cref{fig:timebudget}.

\begin{table}[h]
  \caption{\textbf{Trace statistics per model.} Hypotheses: share of code cells that open with one
  (\cref{fig:hyp}). Paired and bypass: share of hypotheses that propose a paired check or a bypass test,
  and the median paired evaluations run per trial (\cref{fig:tests}). Layer: share of named layers that
  are modified, with chance in brackets (\cref{fig:layers}). Time, passes, calls and tokens: medians per
  trial (\cref{fig:time,fig:steps}); tokens per call in brackets. Cost: median cost per trial at API
  list prices (\cref{fig:cost}).}
  \label{tab:traces}
  \centering\footnotesize
  \setlength{\tabcolsep}{4pt}
  \begin{tabular}{@{}l r r r r r r r r r r@{}}
    \toprule
    Model & \hd{Hypo-}{theses} & \hd{Paired}{checks} & \hd{Paired}{evals} & \hd{Bypass}{tests} & \hd{Layer}{(chance)} & \hd{Time,}{min} & Passes & Calls & \hd{Output tokens}{(per call)} & Cost \\
    \midrule
GPT-6.2 Astra & 100\% & 11.3\% & 8.5 & 5.7\% & 85\% (60\%) & 25.1 & 24 & 141 & 37k (261) & \$4.75 \\
GPT-6.2 Sol & 100\% & 11.0\% & 5 & 4.5\% & 77\% (60\%) & 25.1 & 29 & 165 & 52k (306) & \$1.81 \\
GPT-6 Astra & 100\% & 6.6\% & 14 & 7.6\% & 62\% (42\%) & 25.1 & 23 & 126 & 41k (321) & \$8.57 \\
GPT-6.1 Terra & 99\% & 11.0\% & 18.5 & 9.4\% & 66\% (57\%) & 25.1 & 48 & 327 & 66k (195) & \$1.51 \\
Claude Fable 5.2 & 55\% & 2.7\% & 2 & 3.6\% & 71\% (50\%) & 21.2 & 39 & 109 & 100k (859) & \$6.18 \\
GPT-6.1 Sol & 100\% & 7.2\% & 18.5 & 10.4\% & 53\% (41\%) & 25.1 & 29 & 131 & 45k (336) & \$1.28 \\
Gemini 4 Hydrogen & 86\% & 4.0\% & 2 & 5.6\% & 73\% (52\%) & 25.0 & 15 & 203 & 216k (1037) & \$4.85 \\
Gemini 4 Neon & 83\% & 2.6\% & 1 & 4.1\% & 76\% (50\%) & 25.0 & 11 & 127 & 134k (1128) & \$4.70 \\
Claude Sonnet 5.5 & 12\% & 0.3\% & 2 & 0.7\% & 84\% (63\%) & 17.5 & 47 & 121 & 70k (558) & \$1.42 \\
Gemini 4 Carbon & 87\% & 3.1\% & 2 & 5.6\% & 72\% (51\%) & 25.2 & 8 & 98 & 90k (888) & \$2.29 \\
Claude Opus 5 & 54\% & 3.1\% & 6 & 1.6\% & 76\% (56\%) & 21.9 & 17 & 91 & 69k (746) & \$4.07 \\
Claude Opus 5.5 & 51\% & 1.5\% & 4.5 & 2.4\% & 70\% (49\%) & 17.7 & 39 & 129 & 62k (495) & \$2.73 \\
GLM-5.3 & 77\% & 7.1\% & 12.5 & 2.1\% & 58\% (48\%) & 21.8 & 24 & 183 & 44k (259) & \$0.70 \\
Qwen4 Max & 93\% & 5.8\% & 12.5 & 3.8\% & 57\% (51\%) & 25.0 & 20 & 149 & 49k (318) & \$1.02 \\
GPT-6 Luna & 85\% & 13.1\% & 10 & 5.6\% & 59\% (55\%) & 25.1 & 18 & 118 & 63k (383) & \$0.14 \\
Claude Haiku 5.5 & 70\% & 7.8\% & 3 & 1.7\% & 62\% (44\%) & 10.9 & 17 & 88 & 97k (1086) & \$0.10 \\
DeepSeek V4 Pro & 26\% & 4.7\% & 4 & 0.9\% & 70\% (66\%) & 23.4 & 9 & 107 & 60k (566) & \$0.56 \\
Gemini 3.8 Flash & 62\% & 1.0\% & 5 & 0.7\% & 57\% (46\%) & 23.0 & 12 & 270 & 98k (347) & \$2.45 \\
Mistral Large 4 & 0\% & {\color{ink3}--} & 3 & {\color{ink3}--} & {\color{ink3}--} & 23.5 & 8 & 110 & 57k (499) & \$0.45 \\
    \bottomrule
  \end{tabular}
\end{table}

\begin{table}[h]
  \caption{\textbf{Time budget, per setting} (\cref{fig:timebudget}). Nine problems, two trials each, at
  xhigh effort; the 25-minute row is the setting of the main results.}
  \label{tab:timebudget}
  \centering\small
  \begin{tabular}{@{}l l l r r r r@{}}
    \toprule
    Model & Time & Pass limit & Mean recovery & Median recovery & Trials \geqsev & Median passes \\
    \midrule
Gemini 4 Neon & 5 min & no limit & 11\% & 0\% & 2 of 18 & 3 \\
 & 25 min & 50 passes & 25\% & 20\% & 2 of 18 & 14 \\
 & 120 min & 1000 passes & 83\% & 98\% & 12 of 18 & 93 \\
\addlinespace
GPT-6.2 Sol & 5 min & 1000 passes & 28\% & 18\% & 2 of 18 & 8 \\
 & 25 min & 50 passes & 50\% & 49\% & 4 of 18 & 28 \\
 & 120 min & 1000 passes & 65\% & 71\% & 8 of 18 & 115.5 \\
\addlinespace
Claude Opus 5.5 & 5 min & 1000 passes & 14\% & 5\% & 1 of 18 & 23.5 \\
 & 25 min & 50 passes & 27\% & 17\% & 2 of 18 & 28.5 \\
 & 120 min & 1000 passes & 58\% & 53\% & 7 of 18 & 303 \\
    \bottomrule
  \end{tabular}
\end{table}

\FloatBarrier
\section{Agent Prompts}
\label{app:prompts}

We give both prompts verbatim. They are templates: the placeholders, shown as \ph{NAME}, are filled in
for each problem. The prompts use the internal names of the project: \emph{Repair Weight Gym} is our
harness, the \emph{held-out benchmark} is \btw{}, a \emph{corruption} is a set of bugs and a \emph{link}
is one bug.

\subsection{Bug-Injection Prompt}
\label{app:prompt-injection}

The bug-injection agent receives this prompt as its first message, in a headless Claude Code session in
its own copy of our repository (\cref{sec:bugs}). \ph{MODEL\_CARD} is the network's entry in our list of
candidate networks: the model and the dataset at pinned revisions, the metric, the preprocessing and the
dev and hidden splits, with the conventions shared by all entries. \ph{OUTPUT\_DIR} is a folder outside
the repository, where the agent writes the answer key. The solver limits the prompt quotes (20 minutes,
50 forward passes) are the defaults at the time the problems were built; the evaluation uses the limits
of \cref{sec:solver}.

\begin{promptbox}{Bug-injection prompt}
You are building \textbf{one problem} for a held-out benchmark of neural-network weight repair.
\par
The repository you are in is \emph{Repair Weight Gym}; read \texttt{README.\allowbreak{}md} first. Models have been trained on the gym's corruptions (\texttt{src/\allowbreak{}weightgym/\allowbreak{}corruption.\allowbreak{}py}). This benchmark tests whether they can diagnose and reverse damage they have \textbf{never seen}.
\par
\textbf{What the solver gets.} A solver agent receives your corrupted network and is told only its score on the test set and the score that ``another network with the same architecture'' reaches. It is \textbf{not} told that the network was corrupted. It works for a fixed time (20 minutes by default) with what a Harbor task gives it (the full text is \texttt{src/\allowbreak{}weightgym/\allowbreak{}templates/\allowbreak{}instruction.\allowbreak{}md}; read it):
\begin{itemize}
\item
a persistent Python kernel over the weights: \texttt{load(name)}, \texttt{load\_\allowbreak{}original(name)}, \texttt{save\_\allowbreak{}edit(name,\allowbreak{} array)} (whole-tensor edits), \texttt{drop\_\allowbreak{}edit(name)}, the buffers (read-only), the module list, and numpy, scipy, scikit-learn and pandas for analysis;
\item
the dev data (inputs and targets);
\item
a budget of forward passes (50 by default), each on up to 1,024 dev examples or inputs of its own: \texttt{evaluate} (score, per-example scores, confusions; edited, original or both with a paired difference) and \texttt{read} (the inputs and outputs of any modules, raw or as per-channel statistics).
\end{itemize}
\par
It may not use gradients or any training, may not run the network itself (only through forward passes), and has no internet. It can compute anything else from the weights and the forward-pass results in Python, so \textbf{there are no other tools, and you do not write any}.
\par
\textbf{Your input:} the model described by this card. The card gives the model, the dataset, and exactly how the model must be evaluated. \texttt{conventions} applies to it:
\begin{promptcode}
{{MODEL_CARD}}
\end{promptcode}
\par
\textbf{Your goal:} produce a corrupted version of this model. The corruption must:
\begin{itemize}
\item
\textbf{hurt performance substantially} on the target dataset;
\item
\textbf{change no more than 3\% of the weights}, counted elementwise against the clean model;
\item
\textbf{keep the original information in the weights.} An exact or near-exact repair must exist \emph{without retraining}: the corruption moves, reorders, re-signs or re-maps information, it does not destroy it. Undoing it recovers the clean weights (or a function-equivalent of them) exactly, or very nearly;
\item
\textbf{be structurally difficult to reverse.} Least squares, low-rank fitting, simple rescaling, renormalizing outliers to their peers and similar generic fits must not be sufficient. The repair has to recover a \emph{structure} (which units were moved where, which signs, which pairing, which mapping), not a few continuous values;
\item
\textbf{be discrete, coupled and nonlinear}, not additive. Additive noise destroys information and invites generic fits. Prefer corruptions whose unknowns are discrete objects (a permutation, a sign pattern, an assignment, a choice of block), that interact with each other (they must be found jointly, or in a definite order), and that act nonlinearly on the weights or on the function;
\item
\textbf{use the dev set for identification, not as the solution.} The dev data must help the solver find \emph{what} happened and \emph{where}, but must not directly determine the fix: an edit fitted to the dev examples (for example by regressing a layer's outputs on dev data) should repair the dev score at best partially and generalize poorly, while recovering the network's original structure generalizes to the hidden split almost perfectly;
\item
\textbf{be hard to spot}, so that inspecting the weights is not enough to find the damaged ones, and the obvious ``find the outlier and fix it'' moves do not work;
\item
\textbf{go beyond what the gym already contains.} It must not be one of the existing operators, a re-tuning of one, or a simple combination of them. Design it for \emph{this} model. Study the architecture and the role of each layer, how the layers interact, normalization and its statistics, and what the model relies on for specific classes or inputs. Then exploit that;
\item
\textbf{be solvable by reasoning, not by search}, within the solver's time and forward-pass budget. The difficulty must come from working out \emph{what} happened, not from trying many values (see the next section).
\end{itemize}
\prompthead{3}{Designing for reasoning, not search}
\par
The benchmark measures diagnosis. A problem whose only route to the answer is to guess the kind of corruption and then try many candidate values against the dev score measures search instead, and is not acceptable. For example, 48 independent rotation angles, each found by trying several values, is brute force, however hard it is to spot. A discrete unknown is fine when evidence pins it down (a permutation recovered by a matching computed from activations), not when it has to be enumerated.
\par
\textbf{Requirements:}
\begin{itemize}
\item
\textbf{Evidence points to the mechanism.} There must be observations, available to the solver, that reveal \emph{which} kind of damage happened and \emph{where}: a recognisable fingerprint in the per-class or per-timestep error, a sharp change in what some layer passes on, activations that break a property a healthy layer of this kind always has, and so on. This evidence comes from forward passes, not from the weights alone (see ``No signature in the weights alone'' below). In \texttt{rationale.\allowbreak{}md}, name the observations that lead to each step of the diagnosis.
\item
\textbf{Few, structured unknowns.} Once the mechanism is understood, what remains to be found should be a handful of interpretable unknowns (one permutation, one sign pattern, one assignment per block). Each is recovered by a computation on the weights or on forward-pass results, and the result should be exact, or very close to it. Many independent free parameters are not acceptable.
\item
\textbf{Damage only what the model cannot do without.} A solver that cannot find a link can still cut the damaged part out. So before you design, measure on the clean model, in-process, what removing each candidate site costs: zeroing that layer, block, head, attention or MLP output, or set of channels. Put the corruption where removal is expensive: removing the site must cost more than half of the gap between the clean and the corrupted hidden score you are aiming for. If the model barely notices a site going missing, do not corrupt it.
\item
\textbf{Every link is necessary.} No link may be bypassed by a fallback that needs no diagnosis. For each link, take the corrupted model with every \emph{other} link repaired (only that link still corrupted), apply the best fallback a solver could use without understanding it (within the edit fraction), and measure the hidden recovery. Recovery is on the whole problem's scale, \texttt{(s - s\_\allowbreak{}corrupted) /\allowbreak{} (s\_\allowbreak{}clean - s\_\allowbreak{}corrupted)} with \texttt{s\_\allowbreak{}corrupted} the fully corrupted model, so a solver that finds every other link and patches this one stays below the bar. Try at least:
\begin{itemize}
\item
ablation: zero or remove the damaged component (head, attention block, MLP, layer, channels);
\item
mean or peer replacement: replace the damaged tensor by the mean of its peers;
\item
renormalisation and recalibration: BatchNorm or LayerNorm statistics, gains or weight norms reset to their peers;
\item
simplification: uniform attention, an identity or skipped residual branch;
\item
a closed-form refit on dev activations (least squares, ridge, LDA) of the damaged layer, or of the output layer even when it is not damaged: a classifier head or unembedding rows can be refit from the last hidden states and the dev targets alone.
\end{itemize}
\par
\textbf{Think about the worst case, not the first fallback that comes to mind.} A solver has 50 forward passes to search with: it will try one ablation per block, keep every one that helps on dev, combine them, and add outlier fixes on top. So do the same: for each fallback kind, take the best site and the best \emph{combination} of sites (for example, greedily add the removals that help on dev and keep them together), and mix fallback kinds (ablations plus an outlier fix). Your measured floor is a lower bound on what a solver will find, so leave a margin.
\par
The best fallback, including such combinations, must stay \textbf{below 0.5 hidden recovery} for every link (and on the fully corrupted model). If one passes 0.5, redesign that link; do not just report it. A problem is only solved when every link is found.
\item
\textbf{No signature in the weights alone.} The corrupted tensors by themselves must look healthy. Run the obvious weight-only checks on every corrupted tensor and compare them with the clean model: smoothness and correlation between neighbouring entries, row and column norms, comparison with peer layers, spectra, duplicated or outlier rows. None may stand out. Do not reorder anything that has a natural order a solver can see (pixel or kernel positions, sequence positions, time steps): a smooth filter grid gives such a permutation away. Prefer a symmetry of the network (hidden units, heads, the residual basis) applied on one side of a pair and not on the other: each tensor looks normal alone, and only their mismatch is wrong.
\item
\textbf{The fix cannot be checked from the weights.} If a solver can score candidate fixes without forward passes (for example by how smooth the result is), it will search them for free. The candidates must be told apart only by forward-pass evidence, and there must be many of them (a permutation of 12 heads, not 8 rotations of a grid).
\item
\textbf{Links that only pay together.} Use two or three coupled links. Fixing any one of them alone, with the others still corrupted, must recover \textbf{less than 0.2} (with and without the best fallback on the rest), so that partial progress is not rewarded.
\item
\textbf{Your solution is the solver's path.} Write the solution as a solver would find it, starting blind. If your first weight-only analysis already finds the damage and the fix, the problem is too easy, whatever your time estimate says: redesign it.
\end{itemize}
\par
\textbf{Ways to add depth without adding search.} Combine some of these; they are suggestions, not a checklist:
\begin{itemize}
\item
\textbf{Chain corruptions so that each one hides the next.} The signature of the second only shows up once the first is fixed, and so on. Three to five links, each needing a cycle of diagnose \ensuremath{\to} hypothesise \ensuremath{\to} fix \ensuremath{\to} re-measure, give real depth. Skipping a step leaves the evidence confusing.
\item
\textbf{Make fixes work only together.} Corrupt one site and put a partial counter-corruption downstream that is \emph{not} function-preserving, so that fixing either one alone lowers the score and fixing both restores it. This defeats ``keep what improves dev'' loops and forces a joint hypothesis.
\item
\textbf{Make the first evidence ambiguous.} Choose damage that two or three explanations fit at first, so that the solver has to design a test that tells them apart (a restricted set of examples, the inputs of a specific module, inputs of its own, per-class or per-timestep results).
\item
\textbf{Make the damage depend on the input.} Damage that only hurts some classes, timesteps or inputs is diluted in the overall statistics. The solver must first find the affected subset, then the mechanism.
\item
\textbf{Use decoys sparingly.} A decoy is useful when it competes with the real explanation, not when it is just noise.
\end{itemize}
\par
\textbf{What the solver may know.} This is a hard rule for the evidence your problem relies on. The solver may use only:
\begin{enumerate}
\item
\textbf{the corrupted model}, including any buffers it ships with (such as BatchNorm running statistics, which are part of the model and are never corrupted);
\item
\textbf{the data:} dev inputs and targets, and inputs it makes up;
\item
\textbf{public knowledge:} the architecture, the input preprocessing, and published formulas such as the sinusoidal timestep embedding or the noise schedule.
\end{enumerate}
\par
\textbf{Nothing computed from the clean model} may reach the solver: no clean weights, and no clean activations, statistics, outputs or fingerprints recorded before the corruption. The reference score in the solver's instruction is the only exception. A problem that is only solvable with clean-model information is not acceptable.
\par
\textbf{References within that rule.} Reasoning needs something to compare against. BatchNorm running statistics make the gym's CNN tasks solvable by reasoning. Many models have no buffers, but they still offer references:
\begin{itemize}
\item
\textbf{Known inputs:} fixed input normalisation, the exact formula of a sinusoidal timestep embedding, the known statistics of the noised inputs of a diffusion model at every timestep. From these, the solver can compute what the first layers should receive.
\item
\textbf{Known outputs:} labels for classifiers, the N(0, 1) noise target of a denoiser at every timestep.
\item
\textbf{Peers:} repeated blocks, attention heads within a layer, kernels of the same kind, normalisation gains across depth, and smooth trends such as the growth of the residual stream from block to block.
\item
\textbf{Structure:} invariants that every trained network of this kind satisfies, such as the alignment between a layer's output directions and the next layer's input directions, or the consistency of a residual stream.
\item
\textbf{The data itself:} how much information about the target each layer of the \emph{corrupted} model still carries, measured with forward passes on dev data.
\end{itemize}
\par
These let a solver reason inward from the two ends of the network and across peers. Design the damage so that these references reach it.
\par
\textbf{Work independently.} Do not read other problems' output folders, earlier attempts at this problem, or their transcripts. Each problem must be designed from scratch.
\prompthead{3}{Using the harness}
\par
You have the full harness. Use it to iterate:
\begin{itemize}
\item
\texttt{generate.\allowbreak{}Calibrator(model\_\allowbreak{}id)} keeps the clean model and its splits in memory. Use it to apply candidate changes, score them, and restore the clean model, quickly and in-process.
\item
\texttt{engine.\allowbreak{}WeightSpaceEngine} holds a model the way the solver's backend does, and \texttt{solvers.\allowbreak{}Session} gives the solver's exact API on it in-process (\texttt{load}, \texttt{load\_\allowbreak{}original}, \texttt{save\_\allowbreak{}edit}, \texttt{drop\_\allowbreak{}edit}, \texttt{load\_\allowbreak{}dev}, \texttt{forward}), with the same forward-pass and edit accounting. Load your corrupted weights with \texttt{engine.\allowbreak{}set\_\allowbreak{}start(state\_\allowbreak{}dict)}. Use it to check what the solver sees and to develop your solution.
\end{itemize}
\par
Do not run baseline agents or search baselines, and do not run Harbor: the problem is judged by its design, its solution and your estimates. The exceptions are the removal costs and fallback checks in the requirements above: closed-form computations, run in-process.
\prompthead{3}{Setting up the model}
\par
Before you corrupt anything, make the harness load and evaluate the model exactly as the card says:
\begin{itemize}
\item
Register it as a \textbf{new} model id in \texttt{registry.\allowbreak{}py}, pinned to the card's model and dataset revisions, even if the same checkpoint is already registered: existing entries may use different preprocessing or splits. Don't change existing entries. Implement loading, preprocessing, target construction and the metric in \texttt{models.\allowbreak{}py} following the card (\texttt{tier} tells you how much loader work to expect), and describe the new loader's inputs for the solver in \texttt{engine.\allowbreak{}INPUT\_\allowbreak{}FIELDS} and \texttt{engine.\allowbreak{}INPUT\_\allowbreak{}DESC} (what \texttt{load\_\allowbreak{}dev()} returns and what custom inputs must look like). Don't commit these changes.
\item
Build the dev split (visible to the solver through \texttt{load\_\allowbreak{}dev()} and forward passes) and the hidden split (used only for scoring) exactly as \texttt{eval.\allowbreak{}split} and \texttt{conventions.\allowbreak{}split\_\allowbreak{}procedure} say, including any \texttt{fixed\_\allowbreak{}draws}. Save the index lists and fixed tensors so every evaluation reuses them.
\item
The gym's scores are higher-is-better. If the card's metric is lower-is-better, expose a higher-is-better score to the solver (as the gym does with 1 \ensuremath{-} MSE) and report both.
\item
Run \texttt{prepare.\allowbreak{}py} and compare the clean score with what the card (\texttt{card\_\allowbreak{}reported}, if present) and the model's documentation lead you to expect. If they disagree beyond what the split explains, stop and report the discrepancy instead of corrupting a misconfigured model.
\item
Read the card's \texttt{risks} and say in \texttt{rationale.\allowbreak{}md} how they affect this problem, if they do.
\end{itemize}
\prompthead{3}{Working efficiently}
\par
Long runs are expensive mostly because of waiting, not because of the work itself:
\begin{itemize}
\item
Read only what you need from the source: use \texttt{grep} and line ranges, not whole files.
\item
Print short summaries from your experiments, not full outputs. Write long results to files.
\item
Develop and debug your solution on a few examples, then run the full solution once.
\item
When a job takes minutes, start it once and wait for it with a single blocking wait that ends when it finishes. Do not poll it repeatedly, and do not start several watchers on the same job.
\end{itemize}
\prompthead{3}{Deliverables}
\par
Write everything to \texttt{\{\{OUTPUT\_\allowbreak{}DIR\}\}} (never to git, since it is the answer key):
\begin{enumerate}
\item
\textbf{\texttt{corrupted.\allowbreak{}safetensors}}: the corrupted model, as the full set of editable parameters, with the same names and shapes as \texttt{models.\allowbreak{}editable\_\allowbreak{}parameters} for this model. Buffers (such as BatchNorm running statistics) must remain the clean ones.
\item
\textbf{\texttt{solution/\allowbreak{}}}: a script that repairs the corrupted model the way a solver that understood the problem would, using \textbf{only} what the solver has: a \texttt{solvers.\allowbreak{}Session} on a \texttt{WeightSpaceEngine} started from \texttt{corrupted.\allowbreak{}safetensors} (weights, buffers, dev data, forward passes). It never reads the clean weights, anything computed from the clean model, or this directory, and every decision in it follows from what it measures:
\begin{itemize}
\item
no hard-coded tensor names, indices, orderings or values; the script finds the targets and the unknowns from the weights and the forward-pass results;
\item
each step is a diagnose \ensuremath{\to} hypothesise \ensuremath{\to} fix \ensuremath{\to} re-measure cycle, commented with the evidence it relies on;
\item
no gradients, no training and no running the network outside forward passes, exactly like the solver;
\item
it fits within the problem's forward-pass budget, and the repair is exact or near-exact.
\end{itemize}
\par
Include its log: forward passes used, wall time, number and fraction of weights edited, and the repaired dev and hidden scores. After the solution has run, and separately from it, compare the repaired weights with the clean ones (this check may use them) and report how many of the corrupted elements were restored exactly.
\item
\textbf{\texttt{rationale.\allowbreak{}md}}, containing:
\begin{itemize}
\item
what the corruption does and why it hurts this particular model;
\item
why the original information is still in the weights, and what the exact repair is;
\item
why it is structurally difficult to reverse: why least squares, low-rank fitting, rescaling and other generic fits fall short, and why fitting the dev examples does not generalize while recovering the structure does (argue it; you do not need to run these fits beyond the fallback checks);
\item
\textbf{the fallback checks}: the removal cost of each site you corrupted (measured on the clean model), and for each link, and for the fully corrupted model, every fallback you tried with its dev and hidden scores and hidden recovery. Say why each link cannot be cut out or patched without finding it;
\item
\textbf{the shortcut checks}: the weight-only checks you ran on each corrupted tensor (clean vs corrupted), and the recovery from fixing each link alone;
\item
why it is hard to spot: what a solver sees when it inspects the model, and which natural moves fail or are misleading;
\item
why it is new compared with the gym's corruptions;
\item
the chain of reasoning: for each step, the observation that reveals it, the hypothesis, the fix, and what changes after the fix;
\item
\textbf{the estimated effort} for a strong solver that does not know the answer: the number of tool calls (shell commands and kernel runs), forward passes and minutes it needs, step by step, and where it can go wrong. The default limits are 20 minutes and 50 forward passes; if your estimate does not fit them, say so and recommend limits that do.
\end{itemize}
\item
\textbf{\texttt{meta.\allowbreak{}json}} with these keys (the session tools read them):
\begin{itemize}
\item
\texttt{problem\_\allowbreak{}id}, \texttt{model\_\allowbreak{}id\_\allowbreak{}in\_\allowbreak{}registry}, \texttt{dataset\_\allowbreak{}id\_\allowbreak{}in\_\allowbreak{}registry}, \texttt{card} (the full model card you were given);
\item
\texttt{split}: the dev/hidden split definition (procedure, seed, sizes, index files and their sha256);
\item
\texttt{clean\_\allowbreak{}score}, \texttt{corrupted\_\allowbreak{}score}, \texttt{repaired\_\allowbreak{}score}: each \texttt{\{"dev": .\allowbreak{}.\allowbreak{}.\allowbreak{},\allowbreak{} "hidden": .\allowbreak{}.\allowbreak{}.\allowbreak{}\}} (the repaired one from the solution), plus \texttt{repaired\_\allowbreak{}score.\allowbreak{}hidden\_\allowbreak{}recovery};
\item
\texttt{changed\_\allowbreak{}weights}: \texttt{\{"count": .\allowbreak{}.\allowbreak{}.\allowbreak{},\allowbreak{} "fraction": .\allowbreak{}.\allowbreak{}.\allowbreak{},\allowbreak{} "per\_\allowbreak{}tensor": \{.\allowbreak{}.\allowbreak{}.\allowbreak{}\}\}};
\item
\texttt{solution}: \texttt{\{"forward\_\allowbreak{}passes": .\allowbreak{}.\allowbreak{}.\allowbreak{},\allowbreak{} "wall\_\allowbreak{}sec": .\allowbreak{}.\allowbreak{}.\allowbreak{},\allowbreak{} "exactly\_\allowbreak{}restored\_\allowbreak{}fraction": .\allowbreak{}.\allowbreak{}.\allowbreak{}\}};
\item
\texttt{estimate}: \texttt{\{"tool\_\allowbreak{}calls": .\allowbreak{}.\allowbreak{}.\allowbreak{},\allowbreak{} "forward\_\allowbreak{}passes": .\allowbreak{}.\allowbreak{}.\allowbreak{},\allowbreak{} "minutes": .\allowbreak{}.\allowbreak{}.\allowbreak{}\}} for a strong solver;
\item
\texttt{fallbacks}: \texttt{\{"removal\_\allowbreak{}cost": \{site: clean hidden score with it removed,\allowbreak{} .\allowbreak{}.\allowbreak{}.\allowbreak{}\},\allowbreak{} "fully\_\allowbreak{}corrupted": \{"best": .\allowbreak{}.\allowbreak{}.\allowbreak{},\allowbreak{} "hidden\_\allowbreak{}recovery": .\allowbreak{}.\allowbreak{}.\allowbreak{}\},\allowbreak{} "per\_\allowbreak{}link": [\{"link": .\allowbreak{}.\allowbreak{}.\allowbreak{},\allowbreak{} "best": .\allowbreak{}.\allowbreak{}.\allowbreak{},\allowbreak{} "hidden\_\allowbreak{}recovery": .\allowbreak{}.\allowbreak{}.\allowbreak{},\allowbreak{} "tried": \{fallback: hidden\_\allowbreak{}recovery,\allowbreak{} .\allowbreak{}.\allowbreak{}.\allowbreak{}\}\},\allowbreak{} .\allowbreak{}.\allowbreak{}.\allowbreak{}]\}};
\item
\texttt{shortcuts}: \texttt{\{"weight\_\allowbreak{}only\_\allowbreak{}checks": \{check: "clean vs corrupted result",\allowbreak{} .\allowbreak{}.\allowbreak{}.\allowbreak{}\},\allowbreak{} "single\_\allowbreak{}link\_\allowbreak{}fix": [\{"link": .\allowbreak{}.\allowbreak{}.\allowbreak{},\allowbreak{} "hidden\_\allowbreak{}recovery": .\allowbreak{}.\allowbreak{}.\allowbreak{},\allowbreak{} "with\_\allowbreak{}best\_\allowbreak{}fallback": .\allowbreak{}.\allowbreak{}.\allowbreak{}\},\allowbreak{} .\allowbreak{}.\allowbreak{}.\allowbreak{}]\}};
\item
\texttt{recommended\_\allowbreak{}forward\_\allowbreak{}passes}, \texttt{recommended\_\allowbreak{}edit\_\allowbreak{}fraction} and \texttt{time\_\allowbreak{}limit\_\allowbreak{}min}: the limits the problem should be run with (defaults 50, 0.03 and 20).
\end{itemize}
\end{enumerate}
\par
At the end, give a short summary: the scores, the changed fraction, the solution's forward passes, the best fallback per link, the single-link-fix recoveries, the estimated effort, and anything about the problem you are not sure of. Report only numbers you actually measured.

\end{promptbox}

\subsection{Solver Prompt}
\label{app:prompt-solver}

Every solver receives this instruction as its first message, unchanged across harnesses
(\cref{sec:solver}). The placeholders hold the network's architecture and task, its metric and the
metric's definition, the modified and original scores, the formats of the dev inputs and targets, and
the contents of a forward pass's summary, which depend on the kind of task. The limits are those of
\cref{sec:solver}: \ph{TIME\_MIN} is 25, \ph{MAX\_PASSES} is 50 and \ph{MAX\_EDIT\_PCT} is 3.

\begin{promptbox}{Solver prompt}
You are given a pretrained neural network.
\begin{itemize}
\item
Architecture: \ph{ARCH}
\item
Task: \ph{TASK}
\item
Its \ph{METRIC} on this task's test set: \textbf{\ph{CURRENT}}
\item
Another network with the same architecture reaches \textbf{\ph{REFERENCE}} on the same test set.
\end{itemize}
\par
Your job is to make this network's \ph{METRIC} as high as you can by \textbf{editing its weights directly}: find out what goes wrong inside the network, and change the weights responsible in a way you can justify with evidence.
\par
You have \textbf{\ph{TIME\_MIN} minutes} of working time. The clock stops while the backend computes a forward pass.
\promptrule
\prompthead{2}{Rules}
\par
\textbf{No gradients and no training.} Do not use gradient descent or any other form of training: no backward passes, no automatic differentiation, no optimizers, no numerical gradients from finite differences, and no iterative fitting of the weights to a loss. There are no exceptions to this rule. Every command you run, every code block and every forward pass is recorded and reviewed. \textbf{If you use gradients or training in any form, the attempt is scored 0, whatever its result, and your session is terminated on the spot.}
\par
\textbf{The network runs only through the forward-pass tool.} Do not implement or simulate the network's computation, or a stack of its layers, on any input in your own code. Combining forward-pass results with the weights is analysis and is allowed: for example, applying one layer's weights to the input of that layer recorded by a forward pass.
\par
\textbf{No internet.} There is none, and nothing can be downloaded or installed.
\par
\textbf{Allowed:} any analysis of the weights, of the dev data and of forward-pass results, and any edit of the weights that you compute directly, including closed-form computations (linear algebra, least squares, matchings, sorting).
\prompthead{2}{How you are scored}
\par
When the time is up, the network with the edits saved at that moment is evaluated on the hidden test set. Your score is the share of the gap to the other network that you close:
\begin{promptcode}
score = (final {{METRIC}} - {{CURRENT}}) / ({{REFERENCE}} - {{CURRENT}})
\end{promptcode}
\par
1.0 means you reached \ph{REFERENCE}, 0 means no change, and a negative score means the network got worse. The \ph{METRIC} is defined as: \ph{METRIC\_DESC}. Higher is better.
\par
The dev data you can see comes from the same distribution as the test set but holds different examples. An edit tuned to the dev examples does not carry over to the test set; an edit that fixes how the network computes does.
\prompthead{2}{What you have}
\par
Your working folder is \texttt{\ph{WORK}} (read-write). \ph{PYTHON\_ENV}
\prompthead{3}{1. A Python kernel over the weights: \texttt{py}}
\par
A persistent Python session, like a notebook: variables survive between runs.
\begin{promptcode}
py <<'EOF'
# hypothesis: ...
W = load(NAMES[-1])
print(W.shape, W.std())
EOF
\end{promptcode}
\par
\texttt{py -c 'CODE'} and \texttt{py FILE.\allowbreak{}py} work too. If the last line of a run is an expression, its value is printed. One run may take up to 300 seconds (\texttt{py -{}-timeout S} changes this); a run that takes longer is stopped, and the kernel keeps its variables. Output is cut at 10,000 characters, so print summaries, not whole arrays, and write long results to files. \texttt{py -{}-restart} starts a fresh kernel (all variables are lost).
\par
Defined in advance in the kernel (any other \texttt{python3} gets the same names with \texttt{from wsp import *}):
\par\smallskip\noindent\begin{tabular}{@{}>{\raggedright}p{0.3\linewidth}>{\raggedright\arraybackslash}p{0.66\linewidth}@{}}
\toprule
\textbf{Name} & \textbf{What it is} \\
\midrule
\texttt{np} & numpy \\
\texttt{INFO} & a description of the network and the task: architecture, input format, metric, tensors, buffers, modules \\
\texttt{NAMES} & the names of all parameter tensors: the weights you can edit \\
\texttt{SHAPES} & \texttt{\{name: shape\}} for parameters and buffers \\
\texttt{BUFFERS} & the names of the read-only buffers (for example normalization statistics) \\
\texttt{MODULES} & \texttt{\{module name: module type\}}, in the network's order: the names a forward pass can record \\
\texttt{load(name)} & a float32 array: your saved edit of the tensor if there is one, else the weights as given. Buffers load the same way \\
\texttt{load\_\allowbreak{}original(name)} & the tensor as you were given it, before any edit of yours \\
\texttt{save\_\allowbreak{}edit(name,\allowbreak{} array)} & replace the whole tensor with \texttt{array} (same shape). Returns how many of its elements now differ from the weights as given, and your budgets \\
\texttt{drop\_\allowbreak{}edit(name)} & undo your edit of the tensor (back to the weights as given) \\
\texttt{edited()} & \texttt{\{name: number of changed elements\}} for every tensor you edited \\
\texttt{usage()} & forward passes, edit budget and time: used and remaining \\
\texttt{load\_\allowbreak{}dev()} & the dev data as a dict of arrays (also saved to \texttt{\ph{WORK}/\allowbreak{}dev.\allowbreak{}npz}) \\
\texttt{load\_\allowbreak{}pass(k)} & the arrays of forward pass \texttt{k} (see 3. below), as a dict \\
\bottomrule
\end{tabular}\par\smallskip
\par
\textbf{Edit budget:} at most \ph{MAX\_EDIT\_PCT}\% of all parameters may differ from the weights as given, counted element by element, exactly. An edit that would exceed it is rejected as a whole; dropping or reverting an edit gives its budget back. Edited values must be finite, with \textbar{}w\textbar{} \textless{}= 10,000. Buffers cannot be edited.
\prompthead{3}{2. The dev data}
\par
\ph{DEV\_N} examples with their targets (\texttt{load\_\allowbreak{}dev()}):
\begin{itemize}
\item
inputs: \ph{INPUT\_DESC};
\item
target: \ph{TARGET\_DESC}.
\end{itemize}
\prompthead{3}{3. A forward-pass tool: \ph{MAX\_PASSES} passes in total}
\par
A forward pass runs the network on a trusted server, on up to 1,024 dev examples or inputs of your own, and counts as one pass whatever its size. A request that fails does not count. Forward passes run only through the \texttt{forward} shell command, one request per command (they are not available inside the kernel):
\begin{promptcode}
forward '{"type": "evaluate", "model": "both"}'
\end{promptcode}
\par
It prints the pass's summary; the arrays of pass \texttt{k} are saved to \texttt{\ph{WORK}/\allowbreak{}fwd\_\allowbreak{}\textless{}k\textgreater{}/\allowbreak{}} (one \texttt{.\allowbreak{}npy} file per array, plus \texttt{summary.\allowbreak{}json}), and \texttt{load\_\allowbreak{}pass(k)} loads them in the kernel. Decide what each pass must show before you run it: one pass can record many modules at once. There are two kinds of pass.
\par
\textbf{(a) \texttt{evaluate}: the \ph{METRIC} on dev examples.}
\par
\texttt{\{"type": "evaluate",\allowbreak{} "model": "edited" \textbar{} "original" \textbar{} "both",\allowbreak{} "examples": [dev indices] or \{"start": a,\allowbreak{} "stop": b\}\}}
\begin{itemize}
\item
\texttt{"edited"} (the default) runs your current edits, \texttt{"original"} the weights as given, and \texttt{"both"} runs both on the same examples and reports the paired difference (edited \ensuremath{-} original) with its standard error\ph{PAIRED\_EXTRA}.
\item
\texttt{examples} defaults to the whole dev set.
\item
The summary gives, per model, \ph{EVAL\_SUMMARY}.
\item
Arrays: \texttt{index} (the dev indices), \texttt{\textless{}model\textgreater{}\_\allowbreak{}outputs} (the network's outputs) and \texttt{\textless{}model\textgreater{}\_\allowbreak{}per\_\allowbreak{}example\_\allowbreak{}score} (the \ph{METRIC} is their mean)\ph{EVAL\_ARRAYS}.
\end{itemize}
\par
\textbf{(b) \texttt{read}: values inside the network.}
\par
\texttt{\{"type": "read",\allowbreak{} "model": "edited" \textbar{} "original",\allowbreak{} "examples": .\allowbreak{}.\allowbreak{}.\allowbreak{},\allowbreak{} "modules": [module names],\allowbreak{} "module\_\allowbreak{}inputs": false,\allowbreak{} "outputs": true,\allowbreak{} "reduce": null\}}
\begin{itemize}
\item
Records the output of every listed module as \texttt{\textless{}module\textgreater{}:out} (and its input as \texttt{\textless{}module\textgreater{}:in} with \texttt{"module\_\allowbreak{}inputs": true}), plus the network's \texttt{outputs}.
\item
\texttt{"reduce": "channel\_\allowbreak{}stats"} returns, for each recorded module value, the mean and variance of every channel over examples and positions (\texttt{\textless{}module\textgreater{}:out:mean}, \texttt{\textless{}module\textgreater{}:out:var}; the channels are dim 1 of an [N, C, H, W] map, else the last dim). The raw arrays of one pass are limited to 512 MB, so use it for many modules or large maps.
\item
To read inputs of your own instead of dev examples, save them from the kernel with \texttt{np.\allowbreak{}savez(FILE.\allowbreak{}npz,\allowbreak{} .\allowbreak{}.\allowbreak{}.\allowbreak{})}, one array per input field (\ph{INPUT\_FIELDS}) in the same format and dtype as in \texttt{load\_\allowbreak{}dev()}, and run \texttt{forward '\textless{}request\textgreater{}' -{}-inputs FILE.\allowbreak{}npz}.
\end{itemize}
\prompthead{2}{How the session ends}
\par
After \ph{TIME\_MIN} minutes of working time the session ends. The clock stops while a forward pass computes and runs again as soon as the backend gets any other request, so a pass you wait for costs no time, and anything you send the backend while one runs ends its pause. The edits saved at that moment are your answer: there is no submit step, and nothing that is not saved counts. Once the time is up, edits and forward passes are refused. Save an edit as soon as it helps, and check \texttt{edited()} before the end.
\par
Nobody will reply to you during the session, and ending your turn ends your attempt. Keep working until \texttt{usage()} shows that the time is up. Every \texttt{py} run and every \texttt{forward} command also ends with a line on stderr such as \texttt{[elapsed 7:20 of \{\{TIME\_\allowbreak{}MIN\}\}:00 \textbar{} forward passes 12/\allowbreak{}\{\{MAX\_\allowbreak{}PASSES\}\}]}; trust it rather than your own sense of how much time has passed. If you run out of forward passes, you can still keep going with analysis of the weights alone: the buffers, peer layers and earlier passes' arrays need no passes. Check every edit you save that way against that evidence.
\prompthead{2}{How to work}
\begin{enumerate}
\item
\textbf{Diagnose first.} Look at how the network fails on the dev data and where in the network its computation goes wrong, before you change anything.
\item
\textbf{Localize causally.} Before relying on a component, check that changing it changes the behaviour in the way you predicted.
\item
\textbf{Measure with enough power.} Compare edited and original weights on the same examples (\texttt{"model": "both"}): the paired standard error is much smaller than that of two separate measurements. Keep an edit only if its gain is clearly larger than its standard error.
\item
\textbf{Plan your passes and your time.} \ph{MAX\_PASSES} passes and \ph{TIME\_MIN} minutes go fast; \texttt{usage()} shows both.
\item
\textbf{Work efficiently.} State the hypothesis of each code block in its first comment line; load things once into kernel variables; print compact summaries; keep \texttt{\ph{WORK}/\allowbreak{}notes.\allowbreak{}md} up to date with your findings and your current best edit.
\end{enumerate}

\end{promptbox}

\end{document}